\documentclass[fleqn,10pt]{wlscirep}
\usepackage[T1]{fontenc}
\usepackage{bm}
\usepackage{algorithm}  
\usepackage[algo2e,ruled,vlined,linesnumbered]{algorithm2e}
\usepackage{algpseudocode}  
\usepackage{xspace}         
\usepackage{xcolor}
\usepackage[most]{tcolorbox}
\usepackage{comment}
\usepackage{lineno}
\tcbuselibrary{breakable}
\usepackage{makecell}
\usepackage{multirow}
\usepackage[percent]{overpic}

\newtcolorbox{promptbox}[1][]{
  breakable,
  title=#1,
  colback=gray!5,
  colframe=black,
  colbacktitle=gray!15,
  coltitle=black,
  fonttitle=\bfseries,
  bottomrule=1.5pt,
  toprule=1.5pt,
  leftrule=1pt,
  rightrule=1pt,
  arc=0pt,
  outer arc=0pt,
  enhanced,
  before upper={\parindent=1.5em} 
}

\usepackage[font=small,labelfont=bf]{caption}
\usepackage{subcaption}

\usepackage[most]{tcolorbox}
\tcbuselibrary{listings, minted}

\newcommand{\CodePanelHeight}{0.54\textheight}

\newtcblisting{CodeBox}{
  listing engine=minted,
  minted language=cpp,
  minted options={
    fontsize=\scriptsize,
    baselinestretch=0.9,
    breaklines,
    breakanywhere,
    autogobble,
    tabsize=2
  },
  colback=white,
  colframe=black,
  boxrule=0.4pt,
  left=2mm,right=0.5mm,top=2mm,bottom=2mm,
  sharp corners,
  height=\CodePanelHeight,
  valign=top,
  listing only
}

\usepackage[T1]{fontenc}
\usepackage{listings}
\usepackage{xcolor}
\usepackage{inconsolata} 
\usepackage[most]{tcolorbox} 
\definecolor{codegreen}{rgb}{0,0.6,0}
\definecolor{codegray}{rgb}{0.5,0.5,0.5}
\definecolor{codepurple}{rgb}{0.58,0,0.82}
\definecolor{backcolour}{rgb}{0.95,0.95,0.92}
\definecolor{codeblue}{rgb}{0,0,0.8} 

\lstdefinestyle{whitecpp}{
    language=C++,
    basicstyle=\ttfamily\small,          
    backgroundcolor=\color{white},         
    commentstyle=\color{codegreen},         
    keywordstyle=\color{blue},              
    numberstyle=\tiny\color{lightgray},     
    stringstyle=\color{codepurple},         
    identifierstyle=\color{black},     
    breakatwhitespace=false,
    breaklines=true,
    captionpos=b,
    keepspaces=true,
    numbers=none,
    numbersep=8pt,                       
    showspaces=false,
    showstringspaces=false,
    showtabs=false,
    tabsize=4,
    frame=none,   
    framerule=1pt,                        
    rulecolor=\color{lightgray},          
    escapeinside={\%*}{*)},
    morekeywords={override,final,constexpr,noexcept}, 
}

\newtcblisting{cppcode}[2][]{
    listing only,
    listing engine=listings,
    colback=white,                         
    top=0mm,                                
    bottom=3mm,                             
    left=0mm,                               
    right=0mm,
    boxsep=0pt,
    arc=0pt,                             
    auto outer arc,
    boxrule=0.5pt,                        
    enhanced,
    attach boxed title to top left={        
        xshift=5mm,                         
        yshift*=-\tcboxedtitleheight/2     
    },
    boxed title style={                     
        colback=white,                   
        colframe=white,                    
        boxrule=0pt,                        
        arc=0pt,                           
        left=0pt,right=0pt,               
        top=0pt,bottom=0pt            
    },
    title={                                
        \color{black}                       
        \textbf{#2}                         
    },
    overlay={                               
        \draw[color=lightgray, line width=0.8pt] 
            ([xshift=5mm]title.south west) -- (title.south east); 
        \draw[color=lightgray, line width=0.8pt] 
            (frame.north west) -- (frame.south west); 
    },
    before=\vspace{1em},                  
    #1,
    listing options={style=whitecpp}
}

\title{Discovering heuristics in a complex SAT solver with large language models}

\author[1]{Yiwen Sun}
\author[2]{Furong Ye}
\author[2,3]{Zhihan Chen}
\author[1*]{Ke Wei}
\author[2,4*]{Shaowei Cai}

\affil[1]{School of Data Science, Fudan University,  Shanghai, China}
\affil[2]{Key Laboratory of System Software, Institute of Software, Chinese Academy of Sciences, Beijing, China}
\affil[3]{SeedMath Technology Limited, Beijing, China}
\affil[4]{School of Software, Beihang University, Beijing, China}
\affil[*]{Email: kewei@fudan.edu.cn \& caisw@ios.ac.cn}

\begin{abstract}
The Satisfiability problem (SAT) is fundamental in computational complexity theory and has a wide
range of industrial applications. Optimizing modern SAT solvers in real-world settings is quite challenging due to their intricate architectures. While automatic configuration frameworks have been developed, they rely on manually constrained search spaces. Here we develop AutoModSAT, a framework that uses large language models (LLMs) to automatically optimize SAT solvers. AutoModSAT combines an LLM-compatible modular solver design, unsupervised prompt optimization to diversify generated functions, and an efficient search procedure based on presearch strategy and a $(1+\lambda)$ evolutionary algorithm. Extensive experiments across a wide range of datasets demonstrate that AutoModSAT achieves $40\%$ performance improvement over the baseline solver and $30\%$ improvement over the state-of-the-art solvers. Moreover, AutoModSAT also attains a notable speedup compared to the parameter-tuned alternatives of the state-of-the-art solvers over most of the test datasets. These results demonstrate the potential of LLM-guided heuristic discovery for optimizing complex SAT solvers.

\end{abstract}
\begin{document}
\flushbottom
\maketitle

\thispagestyle{empty}

\section*{Introduction}

The Satisfiability problem (SAT), a fundamental question in logic and computer science, asks whether there exists an assignment to variables in a given Boolean formula to make it true. As the first proven NP-complete problem\cite{npc}, SAT holds immense theoretical significance, and serves as a cornerstone in the computational complexity theory. Efficient solutions to SAT  imply efficient solutions for all the problems in the NP class. Modern SAT solvers, driven by cutting-edge heuristics, are important  for industrial applications, such as in semiconductor manufacturing, cybersecurity, automated reasoning and mission-critical software development~\cite{satsurvey17,satsurvey19}.



After decades of development, modern SAT solvers are equipped with complex heuristics, and industry users need to devote massive efforts to customize different SAT solvers for different real-world applications, which requires labor-intensive modifications by experts. For example, the Electronic Design Automation (EDA) task requires SAT solvers to handle various complex problem instances~\cite{edasurvey}. To address this issue, hyperparameter optimization approaches, also referred to as algorithm configuration, have been developed for automated SAT solver improvement\cite{satautomation, satparameter, satparameter2} so that they can  choose well-performing settings for the given problems. However, the algorithm configuration approaches often suffer from manually-defined search spaces and suboptimal performance improvements.

Large Language Models (LLMs), the recent advance in generative artificial intelligence (AI)~\cite{gpt4, deepseekv3, deepseekr1, qwen, qwen3, llama, gemini}, open up avenues for overcoming the limitations of manually designed frameworks~\cite{algorithmdesign}. After DeepMind's pioneering work on complex mathematical and bin-packing problems~\cite{FunSearch}, LLMs have demonstrated promising applications in automated algorithm design\cite{eoh, reevo, autosat, droc, hsevo, extracting, alphaevolve}.  
However, few studies have been successfully done on optimizing  programs that are as complex as SAT solvers. Firstly, SAT solvers often employ carefully customized data structures to improve computational efficiency, but meanwhile present challenges for automated optimization~\cite{autosat}. In addition,   state-of-the-art SAT solvers, such as Kissat~\cite{kissat} and CaDiCal~\cite{cadical},  have evolved over decades and thus contain hundreds of thousands of tokens, much larger than the context length limit of LLMs.

\begin{figure*}[t]
    \centering
        \includegraphics[width=1\textwidth]{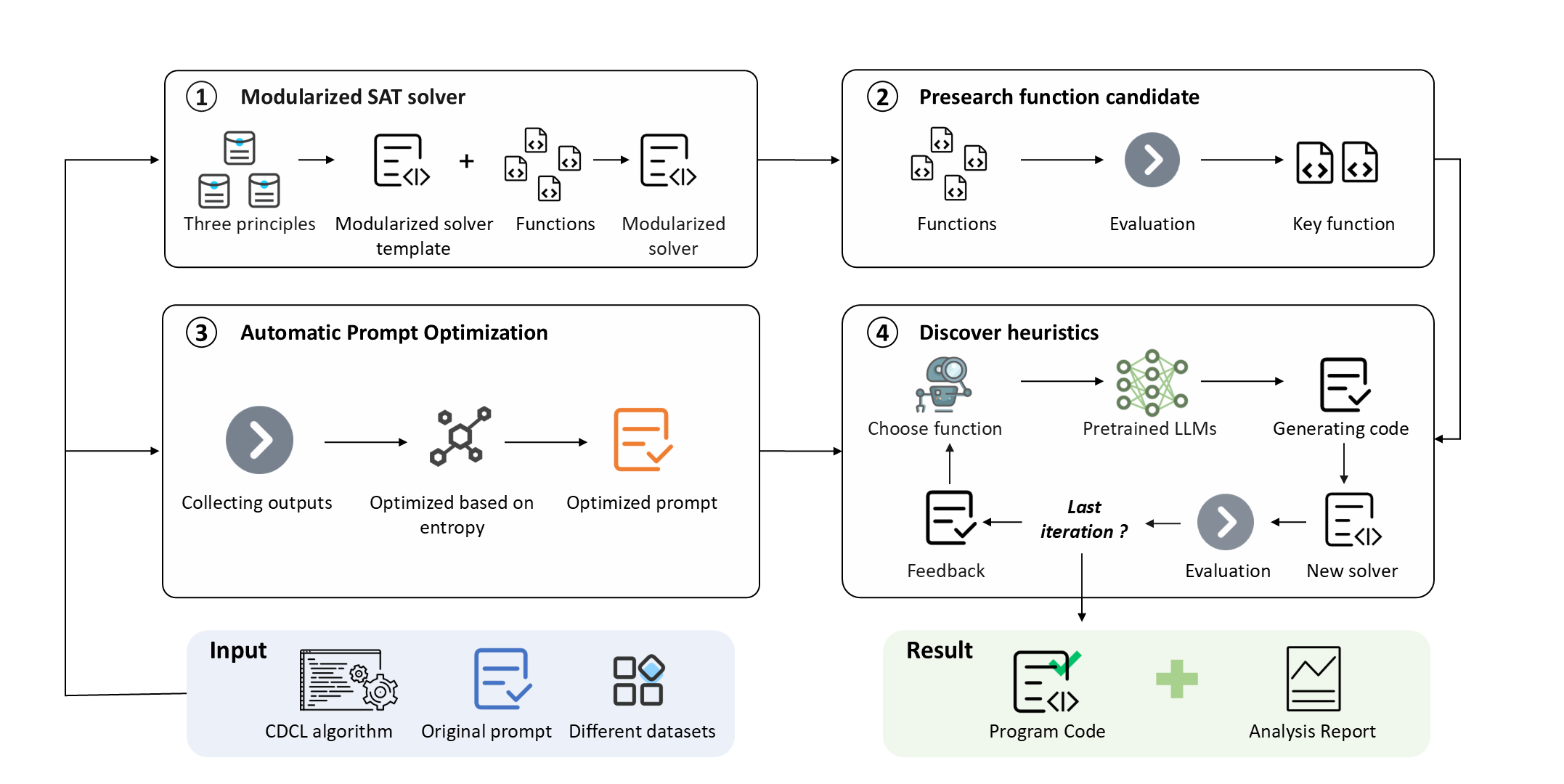}

\caption{\textbf{Overview of AutoModSAT}. This figure displays the workflow of AutoModSAT, which distinguishes itself from existing LLMs methods for algorithm design on challenges in optimizing large-scale, structurally complex solvers. The whole process requires a basic CDCL algorithm, initial prompt, and datasets as input, then involves four core components: (1) a modularizing SAT solver (ModSAT) based on the CDCL algorithm; (2) a presearch strategy to identify function candidates for each dataset; (3) prompt optimization based on entropy; and (4) iterative heuristics discovery. After selecting the function to be optimized, we use LLMs to generate code, and evaluate it on a specific dataset if it can be executed successfully. More effective heuristics are then updated into the solver. }

\label{fig:architecture}
\end{figure*}

To tackle the challenges mentioned above, we propose AutoModSAT, an LLM-driven framework for SAT-solver optimization that is capable of generating competitive heuristics tailored to specific problem instances and thus eliminates repetitive manual tunings across different application scenarios. By leveraging LLMs, AutoModSAT is able to discover competitive heuristics programs beyond the human-designed search space in traditional hyperparameter optimization methods. Furthermore, we develop a well-defined modularized architecture to overcome the token length limitations of LLMs when working on complex solvers. Experimental results show that AutoModSAT achieves better performance across multiple datasets compared to commonly used solvers such as Kissat and CaDiCaL. Overall, AutoModSAT also outperforms the traditional hyperparameter optimization techniques in terms of both solution quality and computational efficiency.

This paper is structured as follows. Firstly, we outline the key components of AutoModSAT, including a modularized SAT solver to optimize, automatic prompt optimization, the presearch strategy and heuristics discovery. Secondly, we evaluate the performance of AutoModSAT across diverse scenario datasets, demonstrating its good performance compared to current SOTA SAT solvers. Then we discuss the insights obtained from AutoModSAT, potential limitations and future research directions. Finally, we provide a detailed description of each component within AutoModSAT, elaborating on its design principles and implementation details.

\section*{Results}
 
\subsection*{AutoModSAT: An automatic heuristic discovery framework for SAT solvers}
AutoModSAT is an LLM-driven framework designed to optimize complex SAT solvers. The whole framework is illustrated in Figure~\ref{fig:architecture}, which comprises four core components. 

\begin{itemize}
    \item \emph{Modularized SAT solver.}
   In order to meet the token length limit and compatibility of LLMs, we introduce ModSAT, a modularized SAT solver with well-designed functions. ModSAT is developed based on three principles, as detailed in the Method section. Instead of leveraging LLMs to generate a complete SAT solver from scratch, seven critical heuristics are defined in ModSAT which form a search space for LLMs to explore. Therefore, AutoModSAT can be expressed as a mapping from a set of heuristics to a solver, given by $ \{h_1, h_2, \ldots, h_7\} \mapsto A$, where each $h_i$ represents a heuristic associated with a function and $A$ is the resulting solver obtained by integrating these functions. 
   The modularized solver enables LLMs to modify heuristics locally without disrupting components in other unrelated functions, facilitating reliable and efficient program generation. 

    \item \emph{Presearch strategy.} A desirable search space not only enables us to generate competitive solvers but can also improve our understanding of the LLM-based search process. An appropriate search space, tailored to the landscape of searching SAT solvers, has been identified through a presearch strategy involving preliminary candidate pruning. 
       \item \emph{Automatic prompt optimization.} Diverse prompts are necessary for LLMs to discover more effective heuristics.
  To this end, we adopt an unsupervised automatic prompt optimization method to increase the diversity of the prompt, which can enhance the performance of the discovered heuristics while reducing labor costs in prompt engineering.
    \item \emph{Heuristics discovery.} After function candidates and prompts are obtained, we leverage three LLM-based agents, a coder, an evaluator, and a repairer to discover effective heuristics based on the  $(1+\lambda)$ EA~\cite{doerr2019theory} search strategy, yielding the final AutoModSAT solver. Extensive experiments across multiple benchmark datasets have demonstrated the superiority of AutoModSAT in efficiency and effectiveness, and provide a direction for future SAT solver development.

\end{itemize}

Overall, AutoModSAT is an LLM-driven optimization framework which can produce domain-specific SAT solvers, and thus can reduce both time and labor costs in real-world applications. It establishes the potential of utilizing LLMs to optimize large-scale solvers that have complex customized data structures and long-text codes. Based on the insights obtained from our practice in this work, it is possible that the proposed LLM-driven optimization framework may also be useful in optimizing other complex solvers.

\begin{figure*}[ht!]
    \centering

    \begin{minipage}{0.5\textwidth}
        \begin{overpic}[width=\textwidth]{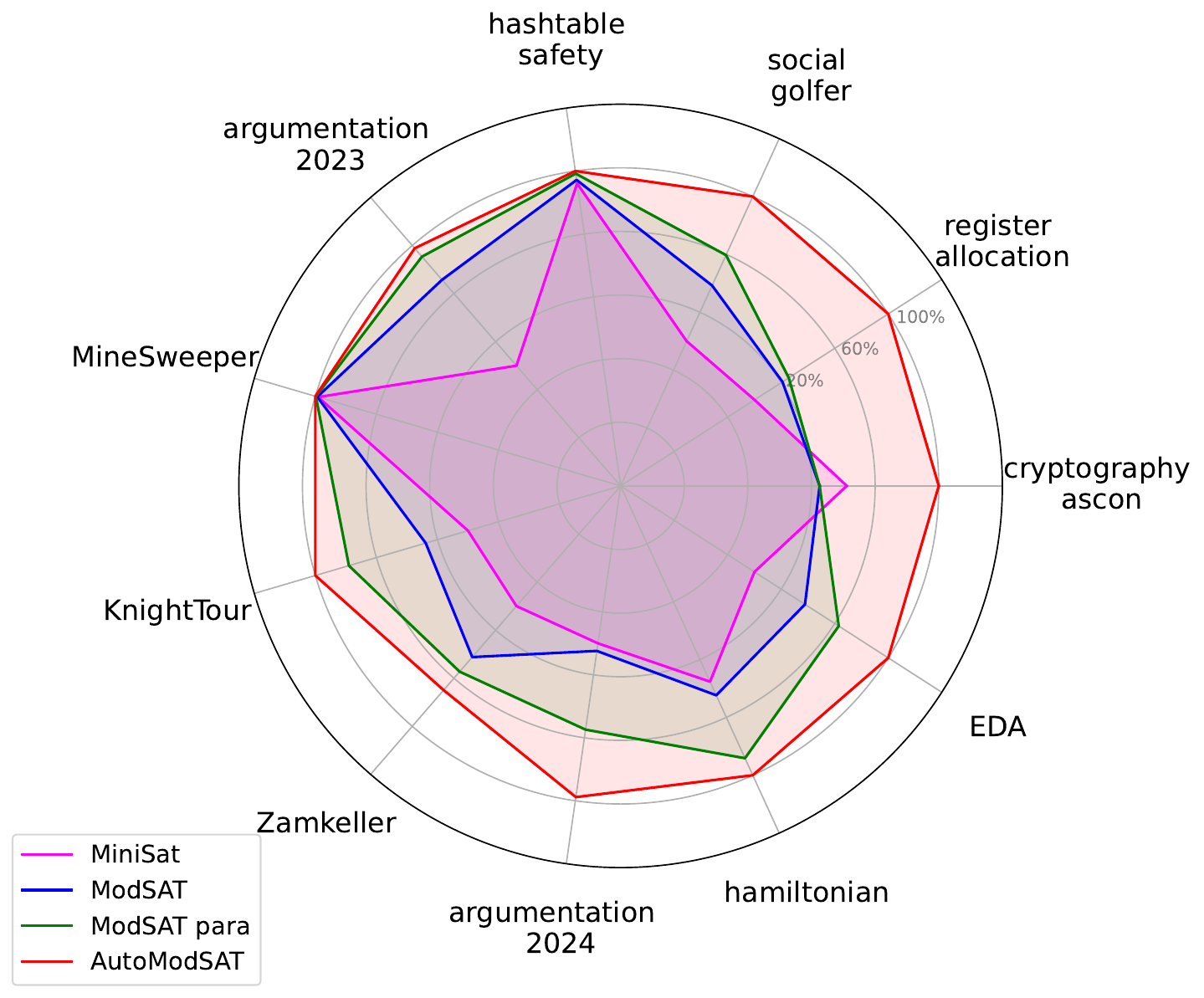}
            \put(2,78){\large\bfseries (a)}
        \end{overpic}
    \end{minipage}\hfill
    \begin{minipage}{0.5\textwidth}
        \begin{overpic}[width=\textwidth]{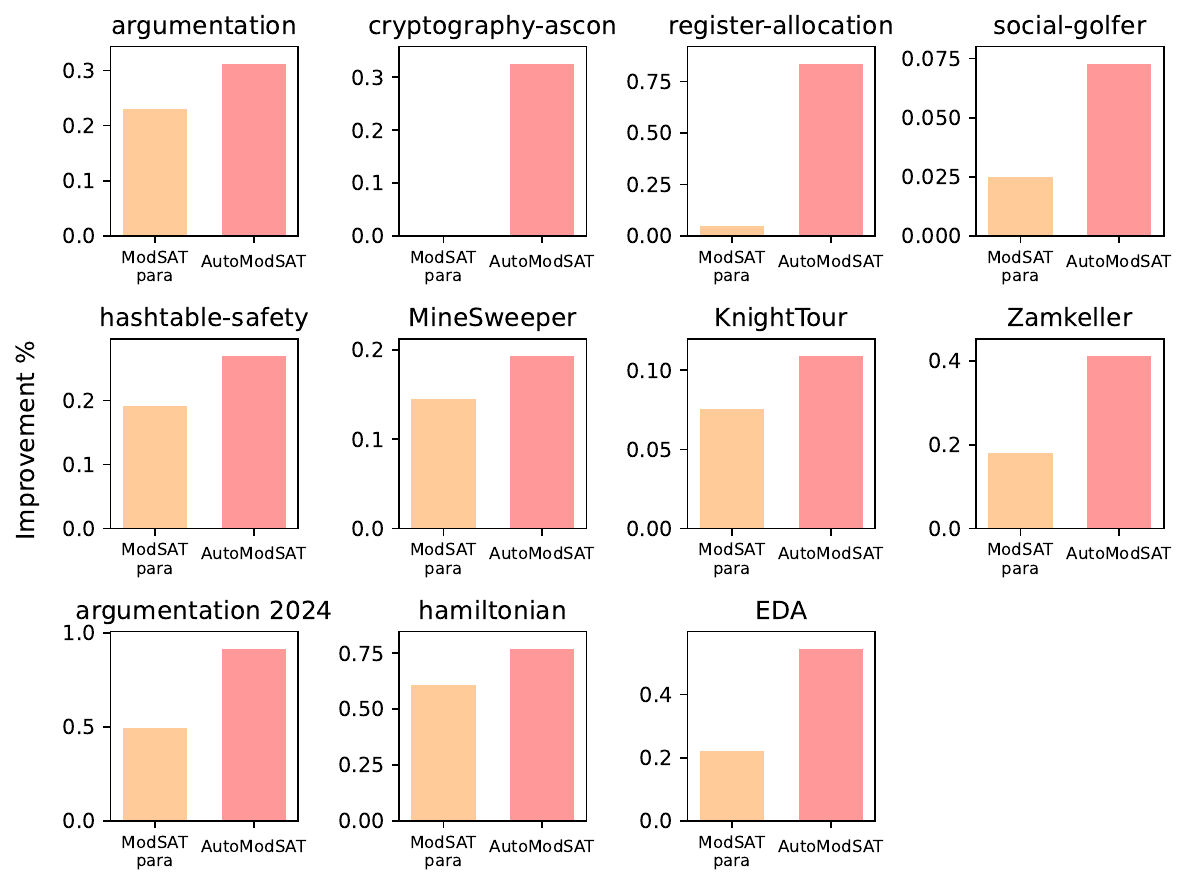}
            \put(2,78){\large\bfseries (b)}
        \end{overpic}
    \end{minipage}

    \begin{minipage}{0.5\textwidth}
        \begin{overpic}[width=\textwidth]{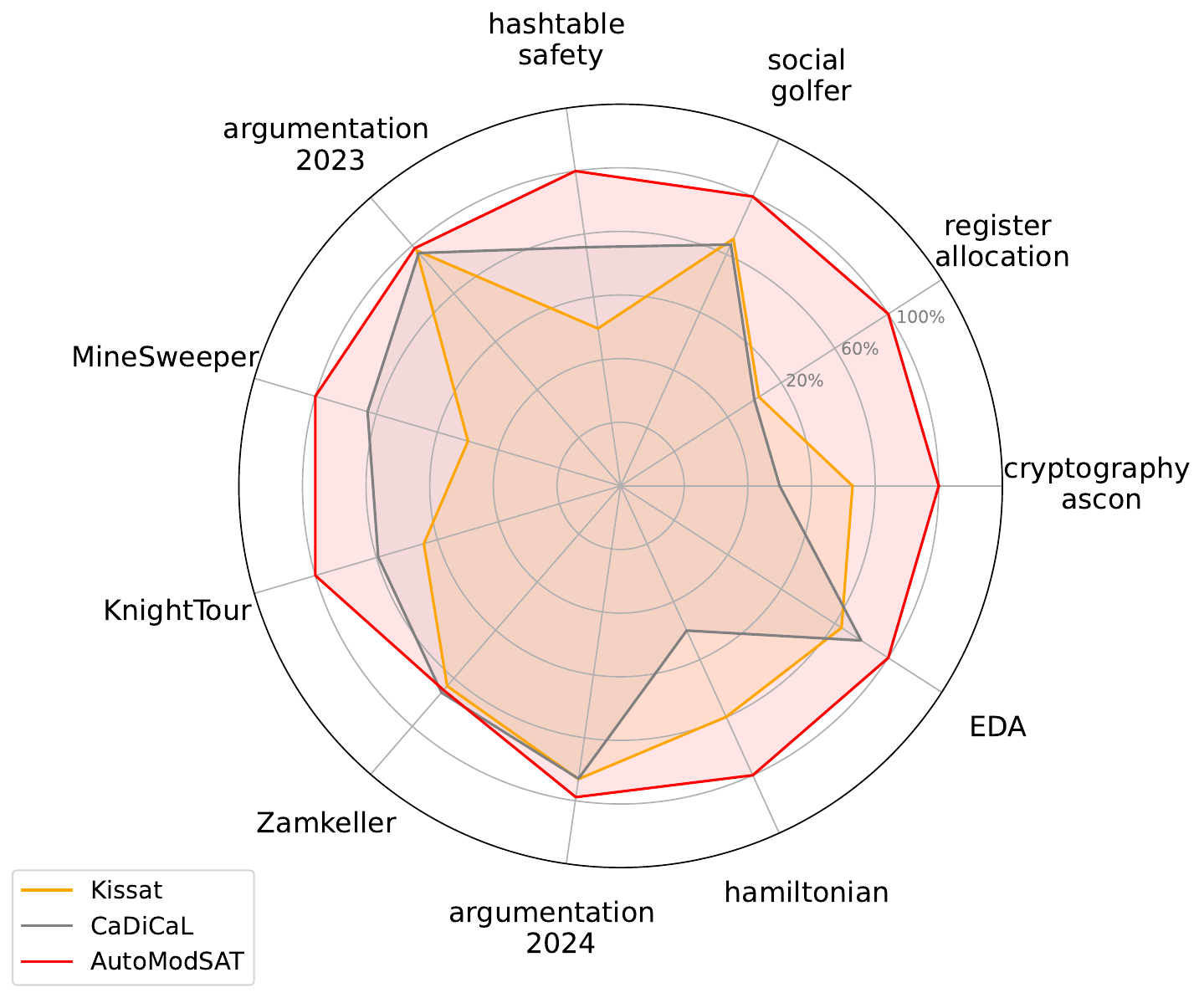}
            \put(2,78){\large\bfseries (c)}
        \end{overpic}
    \end{minipage}\hfill
    \begin{minipage}{0.5\textwidth}
        \begin{overpic}[width=\textwidth]{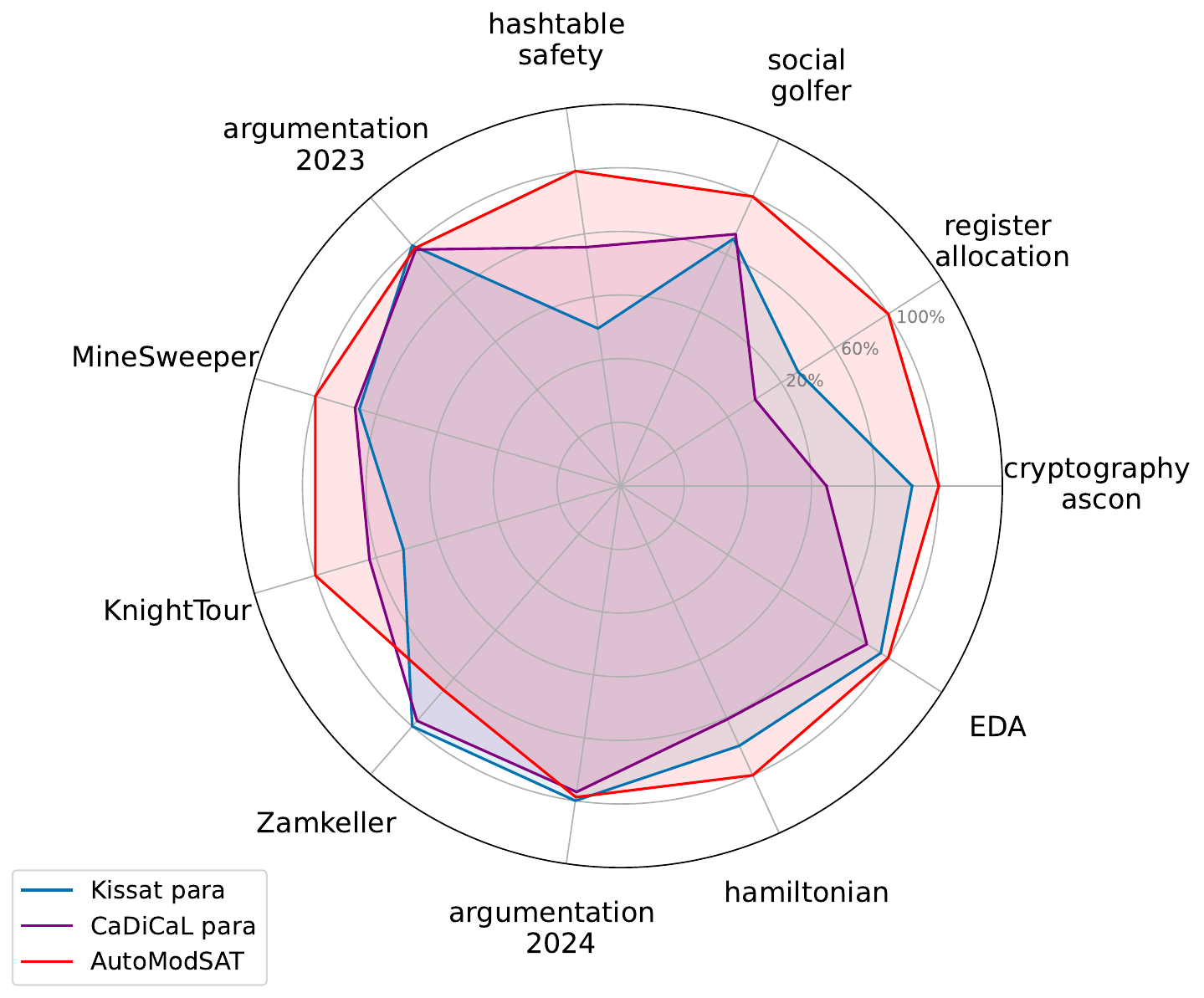}
            \put(2,78){\large\bfseries (d)}
        \end{overpic}
    \end{minipage}

    \vspace{0.5cm}

    \caption{\textbf{PAR-2 over Different Datasets.} 
    \textbf{(a)} This subfigure provides a visualization of PAR-2 in Table~\ref{tab:PAR-2}, which shows the improved performance of AutoModSAT over ModSAT. Here the plotted values are normalized for each dataset by $1 - \frac{curr - min}{2*(max - min)} $, where $curr$ is the PAR-2 corresponding to a solver, and $min$ and $max$ correspond to the minimum and maximum PAR-2 over all the tested solvers. A larger shadow area indicates better performance.
    \textbf{(b)} This subfigure provides a performance comparison between AutoModSAT and parameter-tuned ModSAT on improving the original ModSAT solver, where the vertical axis is the PAR-2 improvement ratio.
    \textbf{(c)} This subfigure compares the performance between AutoModSAT with Kissat and CaDiCaL.
    \textbf{(d)} This subfigure compares the performance of AutoModSAT with parameter tuning variants of Kissat and CaDiCaL.}
    \label{fig:compare}
\end{figure*}

\begin{figure*}[ht!] 

    \begin{minipage}{\textwidth}
        \includegraphics[width=\textwidth]{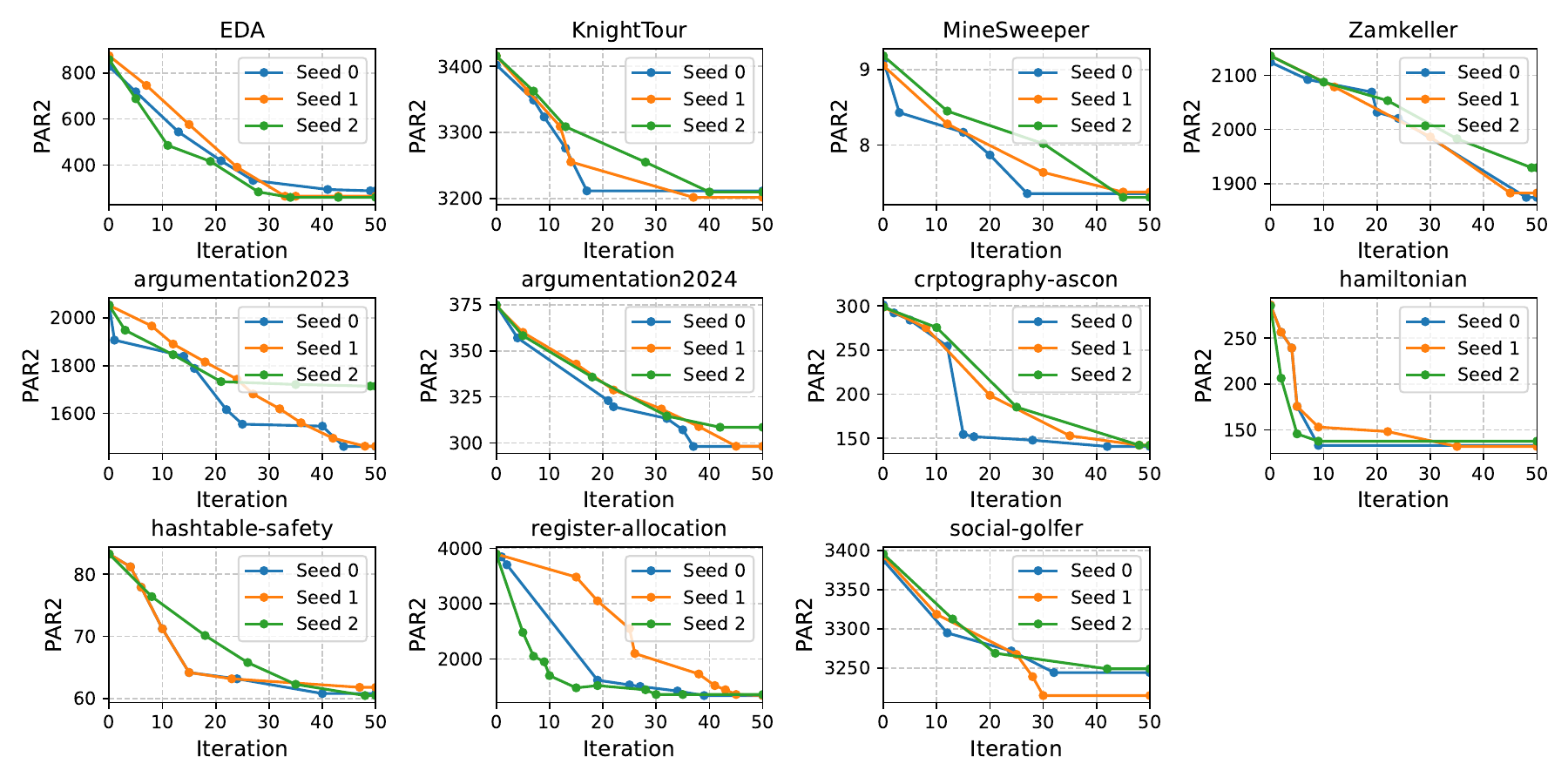}
    \end{minipage}

\caption{\textbf{Iteration Process over Different Datasets.} This figure shows the AutoModSAT optimization trajectory on each dataset for three different runs with different random seeds. The horizontal axis denotes the iteration count and the vertical is PAR-2. }
\label{fig:search_process}
\end{figure*}

\subsection*{Discovery of {effective} heuristics}
We selected $11$ datasets to test the capability of AutoModSAT, consisting of $7$ datasets from the SAT Competition 2023 and 2024~\cite{sat2023, sat2024}, $3$ datasets generated by Picat~\cite{picat}, and another dataset from a real industrial EDA scenario. 
For the selection of SAT Competition datasets, we filter out families with fewer than twenty instances as they are susceptible to overfitting during the iterative optimization process, and select datasets with substantial variations in average solving times in order to comprehensively evaluate a solver's performance across diverse data types. As a result, a total of $7$ families are tested: argumentation\_2023, argumentation\_2024, cryptography-ascon, register-allocation, social-golfer, hashtable-safety and hamiltonian. 
Another $3$ datasets are manually generated by  Picat~\cite{picat}, including KnightTour, MineSweeper, and Zamkeller, which can formulate the constrained satisfied problems into Conjunctive Normal Form (CNF) formulas. 
The last dataset is from an industrial EDA scenario, which contains $50$ industrial instances derived from combinational equivalence checking problems.  The sources and characteristics of all datasets are summarized in Table~\ref{tab:bench}, and more details are presented in Supplementary Section 4.1.

Apart from the classic CDCL-based SAT solver MiniSat~\cite{minisat}, we also compare with the SOTA SAT solvers Kissat~\cite{kissat} and CaDiCal~\cite{cadical}, whose performance {is} fully optimized through hybrid heuristics and complex data structures.  Even though Kissat and CaDiCal have demonstrated superior performance compared to ModSAT, the code complexity and opaque low-level data structures currently prevent effective modifications via LLMs. In addition to the original solvers, we also compare with their parameter-tuning versions in order to fully explore their abilities for different problem instances. More precisely,  we employed SMAC3 (Sequential Model-based Algorithm Configuration)~\cite{smac3} to optimize the parameters of the aforementioned SAT solvers. SMAC3 follows a Bayesian optimization framework and consists of three key steps: 1) define the configuration space, solver-specific parameters and their feasible ranges; 2)  configure SMAC3 using a performance metric (e.g., PAR-2) and execute multiple optimization trials with instance-specific training datasets; 3) the best-found configuration is validated on the testing dataset. This approach systematically addresses parameter interdependencies while mitigating overfitting, leading to performance improvements over default settings. Note that PAR-2 is an evaluation metric that measures the performance of a SAT solver by averaging the runtime for solved instances while penalizing the unsolved instances with a double timeout penalty. Further details about PAR-2 are provided in the Supplementary Section 4.2. It should be emphasized that the focus in this work is the single-core SAT solver and that parallel SAT solving is out of scope, so all solvers are executed in the single-thread mode.

All experiments are conducted on five identical servers, each equipped with two AMD EPYC 7763 CPUs (64 physical cores per CPU) at a base frequency of 2.45\,GHz. All SAT solvers are implemented in C++ with \texttt{g++} 9.4.0, while the interfaces to LLMs and parameter-tuning methods are implemented in Python. The budget $\mathcal{B}$, referring to the number of times requesting new heuristics from LLMs or new parameters from SMAC3, is set to $50$. We only report results obtained from DeepSeek-V3 in this paper due to its competitive performance and meanwhile the low cost in optimizing the SAT solvers, see Supplementary Section 4.4. Different time limits are used for the search process and the final evaluation. During heuristic search, the standard $5000$ seconds timeout in SAT competition is excessively long, so we use the median solving time reported from the original solvers as the timeout. After obtaining the optimized solver, the original $5000$ seconds timeout is used for the final evaluation. This approach significantly reduces the computational overhead while maintaining a meaningful assessment of performance . We adopt PAR-2 as the primary performance metric and additionally report the speedup between solvers, defined as: $\text{speedup} = \frac{v_a - v_b}{\max(v_a, v_b)}$, where $v_a$ is the PAR-2 of method A, $v_b$ is the PAR-2 of method B.

\begin{table*}[t]
\belowrulesep=0pt
\aboverulesep=0pt
\begin{center}
\caption{\textbf{Dataset information}.  Datasets used in this paper are taken from the SAT Competition 2023, 2024 (SC 2023, SC 2024 for short), Picat (a language tool for problem generation), and the industrial EDA scenario. The size of each dataset and statistics on the number of variables and clauses are provided, which clearly show the diversity of the problem instances.}
\vskip 0.04in
\resizebox{0.8\textwidth}{!}{
\begin{tabular}{lcc|cc|cc}
\toprule
dataset            & \makecell{total number \\  of instances} & source & variables mean & variables std & clauses mean & clauses std \\
\midrule
cryptography-ascon           & 20 & SC 2023 & 146,636 & 15,010 & 342,940 & 37,315 \\
register-allocation       & 20  & SC 2023 & 381 & 193 & 5,813 & 5,622 \\
social-golfer          & 20 & SC 2023 & 15,540 & 12,157 & 131,517 & 79,751\\
hashtable-safety              & 20  & SC 2023 & 11,712,548 & 4,874,397 & 53,644,509 & 22,032,387 \\
argumentation 2023             & 20  & SC 2023 & 962 & 190 & 27,960 & 23,034 \\ 
argumentation 2024           & 21  & SC 2024 & 909 & 187 & 35,110 & 29,989  \\
hamiltonian              & 40  & SC 2024 & 511 & 60 & 4,062 & 533 \\
\midrule
MineSweeper              & 88  & Picat & 618,801 & 531,657 & 9,065,224 & 7,909,645  \\
KnightTour           & 56  & Picat & 223,697 & 313,704 & 10,692,883 & 17,482,123  \\
Zamkeller              & 80  & Picat & 24,592 & 22,190 & 310,804 & 335,102 \\
\midrule
EDA              & 50  & Industrial & 1,822 & 1,195 & 6,185 & 4,207 \\

\bottomrule
\label{tab:bench}
\end{tabular}
}
\end{center}
\vskip -0.08in
\end{table*}

\begin{table*}[ht!]
\centering
\scriptsize
\setlength{\tabcolsep}{2pt}
\renewcommand{\arraystretch}{1.12}
\belowrulesep=0pt
\aboverulesep=0pt
\caption{PAR-2 scores over different datasets. Each cell reports the PAR-2 score with the 95\% confidence interval, followed by the number of solved instances in brackets. Lower PAR-2 scores are better, and higher solved-instance counts are better. The results are calculated over ten runs with random seeds from 1 to 10. Bold values indicate the best PAR-2 score and the corresponding solved-instance count for each dataset.}
\label{tab:PAR-2}
\vskip 0.04in

\resizebox{\textwidth}{!}{
\begin{tabular}{lcccccc}
\toprule
Solver
& \makecell{cryptography\\-ascon}
& \makecell{register\\-allocation}
& \makecell{social\\-golfer}
& \makecell{hashtable\\-safety}
& \makecell{argumentation\\2023}
& \makecell{argumentation\\2024} \\
\midrule

MiniSat
& \makecell{193.62 $\pm$ 2.37 \\ (20.00 $\pm$ 0.00)}
& \makecell{8822.50 $\pm$ 177.28 \\ (3.00 $\pm$ 0.00)}
& \makecell{8175.59 $\pm$ 153.24 \\ (4.20 $\pm$ 0.54)}
& \makecell{91.81 $\pm$ 1.39 \\ (20.00 $\pm$ 0.00)}
& \makecell{8649.53 $\pm$ 125.42 \\ (3.00 $\pm$ 0.00)}
& \makecell{2887.18 $\pm$ 37.16 \\ (17.00 $\pm$ 0.00)} \\

ModSAT
& \makecell{207.76 $\pm$ 2.96 \\ (20.00 $\pm$ 0.00)}
& \makecell{7149.15 $\pm$ 143.40 \\ (5.80 $\pm$ 0.32)}
& \makecell{7882.76 $\pm$ 123.35 \\ (4.90 $\pm$ 0.43)}
& \makecell{83.44 $\pm$ 1.71 \\ (20.00 $\pm$ 0.00)}
& \makecell{4704.91 $\pm$ 68.00 \\ (13.00 $\pm$ 0.00)}
& \makecell{2738.42 $\pm$ 53.39 \\ (17.00 $\pm$ 0.00)} \\

ModSAT para
& \makecell{209.49 $\pm$ 3.93 \\ (20.00 $\pm$ 0.00)}
& \makecell{6878.99 $\pm$ 105.95 \\ (6.30 $\pm$ 0.29)}
& \makecell{7690.98 $\pm$ 116.70 \\ (5.00 $\pm$ 0.32)}
& \makecell{67.13 $\pm$ 0.87 \\ (20.00 $\pm$ 0.00)}
& \makecell{3650.53 $\pm$ 64.55 \\ (14.00 $\pm$ 0.00)}
& \makecell{1402.98 $\pm$ 25.15 \\ (19.00 $\pm$ 0.00)} \\

Kissat
& \makecell{191.19 $\pm$ 3.64 \\ (20.00 $\pm$ 0.00)}
& \makecell{8500.85 $\pm$ 250.66 \\ (4.00 $\pm$ 0.36)}
& \makecell{7564.10 $\pm$ 140.31 \\ (5.10 $\pm$ 0.53)}
& \makecell{456.85 $\pm$ 13.34 \\ (20.00 $\pm$ 0.00)}
& \makecell{3317.60 $\pm$ 79.87 \\ (14.00 $\pm$ 0.00)}
& \makecell{539.85 $\pm$ 16.14 \\ (21.00 $\pm$ 0.00)} \\

Kissat para
& \makecell{155.63 $\pm$ 2.86 \\ (20.00 $\pm$ 0.00)}
& \makecell{6303.36 $\pm$ 120.48 \\ (8.20 $\pm$ 0.41)}
& \makecell{7499.44 $\pm$ 161.03 \\ (5.20 $\pm$ 0.61)}
& \makecell{454.99 $\pm$ 11.62 \\ (20.00 $\pm$ 0.00)}
& \makecell{\textbf{3123.08 $\pm$ 68.63} \\ \textbf{(14.00 $\pm$ 0.00)}}
& \makecell{\textbf{169.86 $\pm$ 4.52} \\ \textbf{(21.00 $\pm$ 0.00)}} \\

CaDiCaL
& \makecell{230.20 $\pm$ 5.54 \\ (20.00 $\pm$ 0.00)}
& \makecell{8749.83 $\pm$ 246.48 \\ (3.00 $\pm$ 0.00)}
& \makecell{7516.27 $\pm$ 156.29 \\ (5.20 $\pm$ 0.59)}
& \makecell{251.69 $\pm$ 5.53 \\ (20.00 $\pm$ 0.00)}
& \makecell{3443.29 $\pm$ 95.41 \\ (14.00 $\pm$ 0.00)}
& \makecell{553.38 $\pm$ 16.40 \\ (21.00 $\pm$ 0.00)} \\

CaDiCaL para
& \makecell{203.03 $\pm$ 4.14 \\ (20.00 $\pm$ 0.00)}
& \makecell{8790.38 $\pm$ 243.37 \\ (3.10 $\pm$ 0.22)}
& \makecell{7452.61 $\pm$ 159.01 \\ (5.20 $\pm$ 0.47)}
& \makecell{251.09 $\pm$ 5.46 \\ (20.00 $\pm$ 0.00)}
& \makecell{3269.70 $\pm$ 95.18 \\ (14.00 $\pm$ 0.00)}
& \makecell{322.50 $\pm$ 8.44 \\ (21.00 $\pm$ 0.00)} \\

AutoModSAT
& \makecell{\textbf{140.80 $\pm$ 1.34} \\ \textbf{(20.00 $\pm$ 0.00)}}
& \makecell{\textbf{1178.31 $\pm$ 17.45} \\ \textbf{(18.00 $\pm$ 0.00)}}
& \makecell{\textbf{7253.67 $\pm$ 76.83} \\ \textbf{(5.90 $\pm$ 0.29)}}
& \makecell{\textbf{60.62 $\pm$ 0.65} \\ \textbf{(20.00 $\pm$ 0.00)}}
& \makecell{3228.13 $\pm$ 46.40 \\ (14.00 $\pm$ 0.00)}
& \makecell{232.53 $\pm$ 2.36 \\ (21.00 $\pm$ 0.00)} \\

\bottomrule
\end{tabular}
}

\vspace{1.0em}

\resizebox{0.85\textwidth}{!}{
\begin{tabular}{lccccc}
\toprule
Solver
& hamiltonian
& MineSweeper
& KnightTour
& Zamkeller
& EDA \\
\midrule

MiniSat
& \makecell{652.42 $\pm$ 9.56 \\ (37.40 $\pm$ 0.36)}
& \makecell{9.86 $\pm$ 0.16 \\ (88.00 $\pm$ 0.00)}
& \makecell{9126.73 $\pm$ 116.35 \\ (5.20 $\pm$ 0.32)}
& \makecell{5720.77 $\pm$ 114.48 \\ (37.20 $\pm$ 0.76)}
& \makecell{1102.31 $\pm$ 16.22 \\ (45.00 $\pm$ 0.00)} \\

ModSAT
& \makecell{581.26 $\pm$ 11.21 \\ (37.70 $\pm$ 0.42)}
& \makecell{9.18 $\pm$ 0.19 \\ (88.00 $\pm$ 0.00)}
& \makecell{8766.10 $\pm$ 148.88 \\ (7.10 $\pm$ 0.40)}
& \makecell{3472.66 $\pm$ 72.44 \\ (54.10 $\pm$ 0.47)}
& \makecell{829.92 $\pm$ 17.67 \\ (46.00 $\pm$ 0.00)} \\

ModSAT para
& \makecell{227.87 $\pm$ 4.35 \\ (39.20 $\pm$ 0.16)}
& \makecell{7.81 $\pm$ 0.13 \\ (88.00 $\pm$ 0.00)}
& \makecell{8175.10 $\pm$ 114.79 \\ (11.10 $\pm$ 0.31)}
& \makecell{2860.80 $\pm$ 61.65 \\ (62.20 $\pm$ 0.46)}
& \makecell{648.17 $\pm$ 11.29 \\ (48.00 $\pm$ 0.00)} \\

Kissat
& \makecell{456.84 $\pm$ 9.18 \\ (38.40 $\pm$ 0.34)}
& \makecell{144.89 $\pm$ 3.44 \\ (88.00 $\pm$ 0.00)}
& \makecell{8793.72 $\pm$ 257.04 \\ (7.00 $\pm$ 0.70)}
& \makecell{2218.62 $\pm$ 62.90 \\ (65.10 $\pm$ 0.48)}
& \makecell{626.82 $\pm$ 17.22 \\ (48.00 $\pm$ 0.00)} \\

Kissat para
& \makecell{294.64 $\pm$ 6.86 \\ (39.10 $\pm$ 0.26)}
& \makecell{47.53 $\pm$ 0.99 \\ (88.00 $\pm$ 0.00)}
& \makecell{8521.22 $\pm$ 178.18 \\ (8.30 $\pm$ 0.48)}
& \makecell{\textbf{487.88 $\pm$ 12.83} \\ \textbf{(76.40 $\pm$ 0.13)}}
& \makecell{418.82 $\pm$ 10.15 \\ (49.00 $\pm$ 0.00)} \\

CaDiCaL
& \makecell{931.46 $\pm$ 18.95 \\ (38.90 $\pm$ 0.71)}
& \makecell{54.36 $\pm$ 1.39 \\ (88.00 $\pm$ 0.00)}
& \makecell{8336.58 $\pm$ 217.55 \\ (9.40 $\pm$ 0.59)}
& \makecell{1930.20 $\pm$ 49.42 \\ (67.10 $\pm$ 0.33)}
& \makecell{523.22 $\pm$ 10.53 \\ (49.00 $\pm$ 0.00)} \\

CaDiCaL para
& \makecell{442.38 $\pm$ 9.26 \\ (38.60 $\pm$ 0.35)}
& \makecell{43.65 $\pm$ 1.01 \\ (88.00 $\pm$ 0.00)}
& \makecell{8321.33 $\pm$ 211.94 \\ (10.20 $\pm$ 0.57)}
& \makecell{725.78 $\pm$ 21.45 \\ (74.20 $\pm$ 0.13)}
& \makecell{489.28 $\pm$ 13.83 \\ (49.00 $\pm$ 0.00)} \\

AutoModSAT
& \makecell{\textbf{133.35 $\pm$ 1.31} \\ \textbf{(39.90 $\pm$ 0.18)}}
& \makecell{\textbf{7.32 $\pm$ 0.08} \\ \textbf{(88.00 $\pm$ 0.00)}}
& \makecell{\textbf{7829.74 $\pm$ 66.95} \\ \textbf{(12.70 $\pm$ 0.18)}}
& \makecell{2052.87 $\pm$ 27.06 \\ (66.10 $\pm$ 0.19)}
& \makecell{\textbf{375.28 $\pm$ 4.26} \\ \textbf{(49.00 $\pm$ 0.00)}} \\

\bottomrule
\end{tabular}
}

\end{table*}

The experimental results are presented in Figure~\ref{fig:compare} and Table~\ref{tab:PAR-2}. It can be observed that AutoModSAT achieves over 40\% performance improvement relative to ModSAT, and also outperforms its parameter-tuned version  (denoted ``ModSAT para'') with a 30\% speedup.  
Furthermore, AutoModSAT exhibits substantial improvements over the SOTA solvers Kissat and CaDiCaL on each dataset, achieving an average performance gain exceeding 30\%.  Compared with the parameter-tuned Kissat and CaDiCaL (denoted  ``Kissat para'' and ``CaDiCaL para''), AutoModSAT achieves the shortest PAR-2 score while matching or exceeding the solved instance count in 8 out  of the 11 datasets, including cryptography-ascon, register-allocation, social-golfer, hashtable-safety, hamiltonian, EDA, Minesweeper and KnightTour. Note that  ``Kissat para'' achieves the best performance on Zamkeller, substantially better than AutoModSAT. The potential reason for the huge improvement from Kissat to ``Kissat para'' is investigated in Supplementary Section 5.3.

We have also conducted the in-domain and cross-domain generalization tests on the solvers produced by AutoModSAT. The details and the results are presented in Supplementary Section 4.6. As an automation framework of domain-specific solvers, it is expected that the solvers should generalize well for test data in the same domain, while it will be difficult for the solvers to generalize across domains, which matches our observations from the experimental results. 

In addition, we show the iteration process of the heuristic discovery in Figure~\ref{fig:search_process}, where three experiments are conducted independently. It can be seen that AutoModSAT gradually produces the effective heuristics for every dataset. 

Finally, we report the computational time (average runtime with 95\% confidence intervals) required in the heuristic discovery process, which involves three LLM agents. The LLM coder requires 21.05 $\pm$ 0.55 seconds, the LLM evaluator requires 20.00 $\pm$ 0.91 seconds, and the LLM repairer requires 20.37 $\pm$ 0.51 seconds. The runtime is fairly balanced across agents, with each agent typically costing around 20 seconds. Therefore, one full iteration usually takes about 1--2 minutes (and can increase when repeated failures trigger additional repair cycles). More details, as well as a comparison across different LLMs, are provided in Supplementary Table 6.

\subsection*{Investigating the Case on Zamkeller}
To understand why ``Kissat para'' is superior on Zamkeller, we conduct two controlled ablation studies.
\begin{itemize}
    \item \textbf{One-factor ablation.} Starting from default configuration of Kissat, we enable exactly one tuned parameter at a time, while keeping all other parameters at default values. This isolates the marginal effect of each parameter.
    
    \item \textbf{Two-factor combination.} From the one-factor results, we identify the two parameters with the largest positive impact on Zamkeller. We then evaluate the combination: Kissat default + \{eliminate=0, simplify=0\}, and compare it against ``Kissat para''.

\end{itemize}

The PAR-2 scores for every setting is presented in Table~\ref{tab:zamkeller_ablation}. It is clear that eliminate=0 together with simplify=0 explain the majority of the performance gain of  ``Kissat para'' over Kissat on Zamkeller. This suggests that aggressive preprocessing is detrimental on Zamkeller under the default Kissat configuration. In contrast, the current version of AutoModSAT only searches over 7 functions and does not control top-level preprocessing switches such as eliminate and simplify. Note that if we disable these two parameters manually, the performance of AutoModSAT can indeed be further improved on Zamkeller. However, it is still far from competitive compared with ``Kissat para''  since the baseline solver ModSAT is relatively weak on this dataset.

\begin{table}[t]
\centering
\small
\caption{\textbf{Ablation studies of Kissat options for Zamkeller.}
\textbf{One-factor ablation} starts from \textsc{Kissat default} and sets exactly one parameter to the tuned value.
Two-factor evaluates the combination eliminate=0+simplify=0.
The PAR-2 (lower is better) with 5000s timeout is reported.
}
\label{tab:zamkeller_ablation}
\begin{tabular}{l l r}
\toprule
Setting & Change from default & PAR-2 \\
\midrule
Kissat default & -- & 2229.90 \\
Kissat para   & (full tuned config) & 486.75 \\
\midrule
\multicolumn{3}{l}{One-factor ablation: $\textsc{Kissat default} + \{\text{one option}\}$} \\
\midrule
Ablate-eliminate     & eliminate=0      & 560.59 \\
Ablate-phase         & phase=0          & 1999.49 \\
Ablate-probe         & probe=0          & 2139.03 \\
Ablate-reduceint     & reduceint=1e4    & 1739.60 \\
Ablate-rephaseint    & rephaseint=1e4   & 1822.06 \\
Ablate-restartmargin & restartmargin=25 & 1850.55 \\
Ablate-simplify      & simplify=0       & 495.96 \\
Ablate-stable        & stable=2         & 1874.50 \\
Ablate-target        & target=0         & 4138.99 \\
Ablate-tier1         & tier1=4          & 1829.36 \\
Ablate-tier2         & tier2=8          & 1873.41 \\
\midrule
\multicolumn{3}{l}{Two-factor combination: \textsc{Kissat default} + \{eliminate=0,\,simplify=0\}} \\
\midrule
Ablate-eliminate-simplify & eliminate=0, simplify=0 & 488.25 \\
\bottomrule
\end{tabular}
\end{table}

\subsection*{Analysis of discovered heuristics}
AutoModSAT is capable of discovering effective heuristics, and we analyze two of them here: restart function and varBumpActivity as presented in Figure~\ref{fig:code}. More examples are given in Supplementary Figures 9-16. 

The heuristic varBumpActivity increases a variable's activity score to prioritize it for future branching decisions. The generated varBumpActivity function introduces several key enhancements over the original one. For example, it scales the activity increments based on the current decision level of the solver ($1.0 + 0.1 * decisionLevel()$), prioritizing variables involved in recent conflicts. This improves search efficiency by dynamically focusing heuristic guidance on newer decisions. Then, the rescaling mechanism uses a larger threshold ($1e100$ vs. $1e50$) and a smaller scaling factor ($1e-100$), which preserves relative activity differences more effectively while preventing floating-point overflow. Additionally, a minimum activity floor ($1e-100$) ensures no variable becomes entirely inactive, maintaining heuristic relevance. Finally, the heap update logic is optimized to avoid redundant operations by only triggering a decrease when the variable’s activity surpasses the top elements of the heap. These changes reduce computational overhead while maintaining the  ability to adaptively prioritize variables.

The heuristic restart\_function can help the solver escape the local minimum. The generated restart\_function introduces several key enhancements over the original implementation. For example, it implements a dynamic restart strategy using moving averages of Literal Block Distance (LBD) scores, maintaining both fast ($90\%$ historical weight) and slow ($99\%$ historical weight) averages. This enables adaptive decision-making: when recent conflicts become more complex (fast\_avg/slow\_avg ratio > 1.2), it performs a full restart; for moderate difficulty (ratio >1.0), it preserves some learned clauses through partial restart; and for easier conflicts, it minimizes backtracking. Moreover, it also introduces a periodic database reduction to help maintain memory efficiency. These improvements create a restart strategy that adapts to problem difficulty, and balances exploration and exploitation of learned clauses. It is worth noting that there are  a few constants in the adaptive backtracking strategy introduced by LLMs. One should be careful about them since their values are simply  the outputs of the LLMs and lack interpretability. In addition, there are static local variables in the LLM-generated restart\_function. Even though this does not affect our experiments (one solver in a single-process), it would be problematic for parallel SAT solvers. In order to further address this issue, a potential solution is to instantiate the solver multiple times in a single process and record the standard statistics (e.g, conflicts, starts, decisions, propagations) of each instantiation and  see whether they are identical. If not, we can trace back where the problem arises either manually or by LLMs.

\begin{figure}[t]
\centering

\includegraphics[width=1\textwidth]{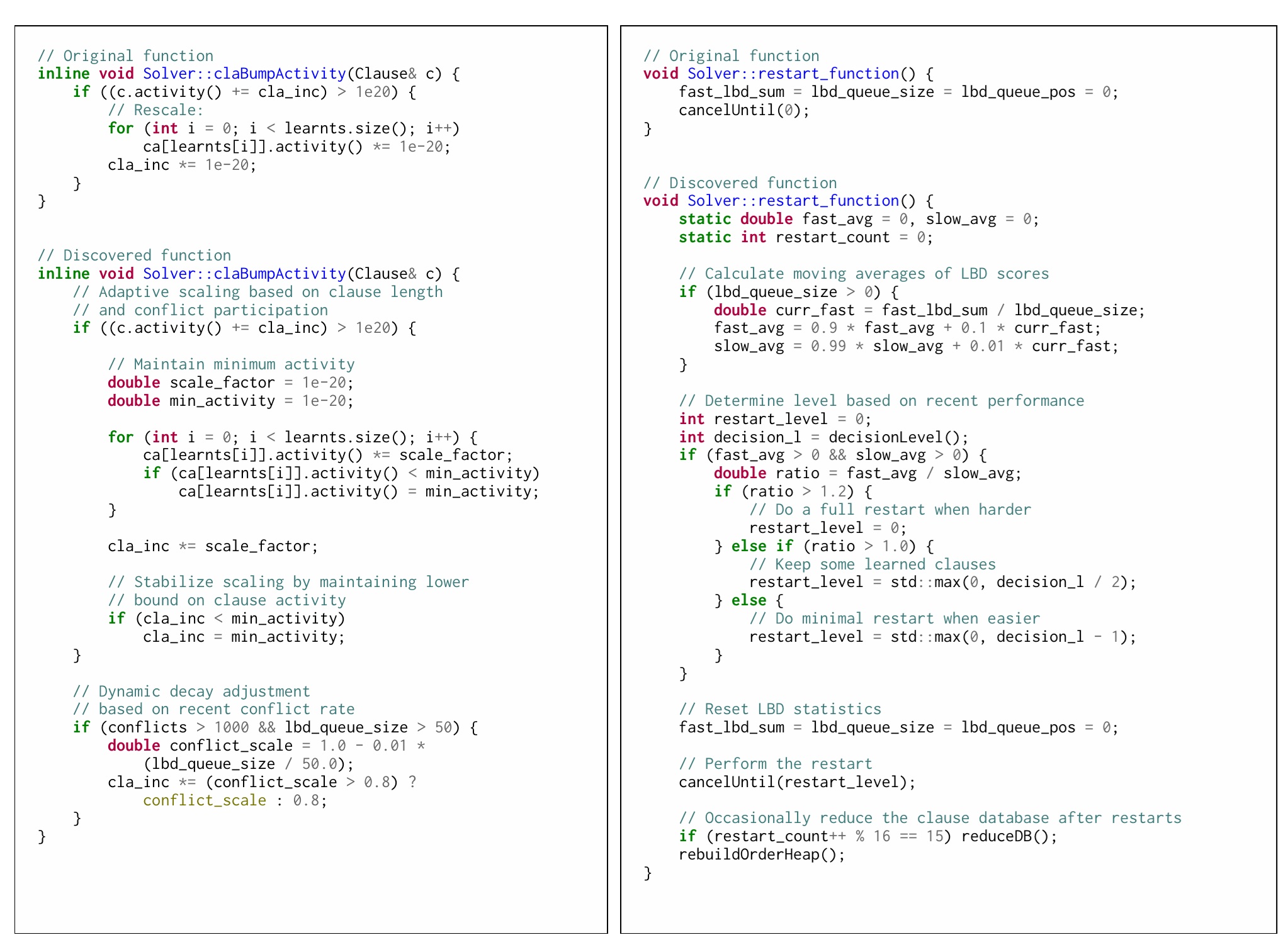}

\caption{This figure shows two effective LLM-generated heuristic functions for SAT solvers and their original counterparts. (Left) The claBumpActivity function innovatively scales activity increments based on the solver's current decision level. Both functions introduce unprecedented heuristic rules not previously proposed in existing solvers, demonstrating exceptional performance on EDA datasets. (Right) The restart function implements a dynamic restart strategy using moving averages of Literal Block Distance (LBD) scores. }
\label{fig:code}
\end{figure}

\section*{Discussion}
The experimental results demonstrate that AutoModSAT achieves substantial performance improvements over both the baseline modularized solver (ModSAT) and existing state-of-the-art SAT solvers, and also attains a notable speedup compared to the parameter-tuned alternatives of the state-of-the-art solvers  across multiple benchmark datasets. These gains demonstrate the hypothesis that LLMs can be leveraged to explore and optimize complex solvers. 

A key innovation of AutoModSAT lies in the solver modularization principles. Traditional SAT solvers have deeply entangled components and large, legacy codebases that are difficult to optimize. By enforcing the three modularization principles and decoupled heuristics, it is easier for LLMs to reason about the solver's behavior at a higher level of abstraction. Although this modularized solver is not as efficient as the well-established solvers such as CaDiCal or Kissat, it provides the necessary foundation for heuristic discovery, which ultimately leads to a more efficient new solver.
The results also highlight the importance of automatic prompt optimization and presearch strategies. Without these mechanisms, the LLM-generated modifications tend to produce suboptimal heuristics. Entropy-based prompt optimization improves both the stability and diversity of the candidate heuristics, while the presearch stage effectively prunes the search space. Together, these components make the $(1+\lambda)$ EA optimization process more sample-efficient, enabling AutoModSAT to discover highly efficient solver variants within practical computational budgets. The ablation studies to investigate the effect of each component are provided in Supplementary Section 4.5.

Note that recent LLM-driven algorithm design methods (e.g., FunSearch, EoH, AlphaEvolve)~\cite{FunSearch,eoh,alphaevolve} show that LLMs can be used as search tools over code or heuristics generation. AutoModSAT is related in spirit, but it focuses on a practical gap that is less emphasized in prior work: making LLM-based optimization work reliably for  SAT solvers that are large, entangled, and hard to automatically optimize reliably. In contrast to the aforementioned LLM-based algorithm design methods, AutoModSAT does not attempt to generate the entire algorithm from scratch due to the complexity of the SAT solvers as well as the quest for high performance.
Compared to recent LLM-based AutoSAT~\cite{autosat}, AutoModSAT achieves much superior performance by introducing several new components, such as an LLM-friendly modularized solver design, automatic prompt optimization to increase code-generation diversity, and a presearch strategy to narrow down the effective search space.

Although these comparisons highlight the advantages of AutoModSAT over prior LLM-based algorithm design and SAT-solver optimization methods, several limitations remain and suggest directions for future work. Firstly, the baseline modularized solver can still be further optimized to provide a stronger starting point. AutoModSAT is not uniformly best across all datasets. While on families where ModSAT is relatively weak (e.g., Zamkeller), the parameter tuning variants of Kissat and CaDiCaL are still superior. Note that it may also be possible to further improve the performance of AutoModSAT by introducing more heuristic functions. Secondly, we would like to emphasize that even though the LLMs are used to discover the heuristics automatically, the entire process is not fully automatic. For example, it requires a human-in-the-loop cycle where an expert prepares the base solver. Lastly, the feedback integration between the search evaluation and the code generation is still heuristic. It is likely that more principled reinforcement or gradient-based feedback loops could make the optimization process more data-efficient.

\section*{Methods}
\subsection* {A brief introduction of SAT}
The SAT problem is defined as {follows}. Let $V = \{x_1, x_2, \dots, x_n\}$ be a set of Boolean variables. A literal is either a variable $x \in V$ or its negation $\neg x$. A clause is a disjunction of literals. A conjunctive normal form (CNF) formula $F = C_1 \land C_2 \land \dots \land C_m$ is a conjunction of clauses. The goal of SAT is to determine whether there exists an assignment of binary values (True or False) to the variables in $V$ such that the value of $F$ is True.

The naive approach for SAT is to  exhaustively enumerate all possible assignments. A complete SAT {solver implements} the search in {a} systematic way, and determines if there exists at least one assignment that satisfies $F$ (proving $F$ is satisfiable). If no satisfying assignment exists, the solver conclusively proves $F$ is unsatisfiable. Within this framework, the strategies for branching (choosing which variable and what value to assign next) become critically important. Key techniques to enhance search efficiency include restart strategies, rephase strategies, and activity-based heuristics (often implemented by bumping variable or clause activity scores).

Conflict-Driven Clause Learning (CDCL)~\cite{cdcl} is the most common approach for complete SAT algorithms, which dominates in modern high-performance SAT solvers. The core innovation is conflict analysis: when a conflict occurs, the solver analyzes it and derives a new clause that explains the inconsistency. These derived clauses, called learnt clauses, enable the solver to prune large portions of the search space, significantly improving efficiency. However, an excessive number of learnt clauses can degrade the unit propagation speed and consume excessive memory. Consequently, identifying high-quality learnt clauses and reducing their quantity are also essential for maintaining solver performance. Therefore, achieving competitive performance in modern SAT solvers requires not only the theoretical advances, but also a series of engineering innovations, including efficient data structures and heuristic policies to realize these ideas effectively.

\begin{algorithm2e}[ht!]
\caption{ModSAT}
\label{alg:modsat}
\textbf{Input:} A CNF $F$ of SAT instance\;
\textbf{Initialization: } decision level $d \leftarrow 0$, current assignment of variables $\mathcal{X} \leftarrow \emptyset$\;

\While{\textbf{True}}{
$\mathcal{X} \leftarrow \textbf{Unit Propagation}(F,\mathcal{X})$\;
\eIf{Conflicts are detected in $\mathcal{X}$}{
\eIf{$d == 0$}{return \textsc{UNSAT}}
{$\mathcal{C}_{conflict}, d_{backtrack} \leftarrow \textbf{Analyze Conflict}(F,\mathcal{X})$ \;
$C_{learned} \leftarrow \textbf{Learn Clause}(\mathcal{C}_{conflict},\mathcal{X})$ \;
$F \leftarrow F \land C_{learned}$\;
$\mathcal{X} \leftarrow \textbf{Backtrack}(\mathcal{X},d_{backtrack})$\;
$d \leftarrow d_{backtrack}$\;
}

\eIf{Restart condition}{$d \leftarrow \textbf{Restart}(d)$}{ continue }
\eIf{Rephase condition}{$\mathcal{X} \leftarrow \textbf{Rephase}(\mathcal{X})$}{ continue }
\eIf{Reduce condition}{$C_{learned} \leftarrow \textbf{Reduce}(C_{learned})$}{ continue }
}
{
\eIf{All variables are assigned with a value}{return \textsc{SAT}}
{
$d \leftarrow d + 1$\;
$\mathcal{X} \leftarrow \textbf{Make Decision}(\mathcal{X})$
}
}
}
\end{algorithm2e}

\subsection*{ModSAT: A Modularized SAT solver}
The current architecture of complex SAT solvers has extensive functions, often resulting in poor modularity. This hinders the direct application of LLMs to solver optimization. Therefore, we develop a modularized solver, ModSAT (see Algorithm~\ref{alg:modsat}), by augmenting MiniSat~\cite{minisat} with rephasing heuristics and then modularizing it according to three principles that will be detailed later. ModSAT follows the basic CDCL framework~\cite{cdcl},
which usually initiates with an empty set of partial assignments (line 2). The Unit Propagation (UP), also called Boolean Constraint Propagation, assigns values to variable in clauses which has only one variable. UP can always make a clause satisfied, and this operation will repeat until no more UP is possible (line 4). If no conflicts are detected in $\mathcal{X}$ during the Decision Detection phase, the algorithm will select a variable to assign a value (line 19). This Make Decision step usually contains a few heuristics. When conflicts are detected (line 5), Analyze Conflict will identify the conflicted clauses (line 9), and a newly learnt clause will be derived based on the clause (i.e., the current partial assignment) of conflicts (line 10). Afterwards, the algorithm backtracks to an earlier decision level (lines 12-13). Once all variables are assigned and no conflict is detected, the algorithm obtains a satisfied assignment (line 28); otherwise, detecting conflicts at the decision level $0$ indicates that the given CNF is unsatisfiable (line 7).

There are several important heuristics in ModSAT. For instance, \emph{reduce} heuristics identify and remove the learnt clauses by controlling the size of the tracking list. In addition, \emph{bump var heuristics} are usually incorporated in the Analyze Conflict step, which affect the choice of variables in the Make Decision function. While the order of choosing variables can determine the search path of branching, 
\emph{rephase} heuristics can control the polarity in {variables} to be selected. Also, \emph{restart} heuristics may abandon the current search path, allowing algorithms to explore possibly easier search regions.
In particular, we have defined seven functions which are independently implemented: \textbf{restart function}, which manages restart heuristics; \textbf{restart condition}, which determines when to execute restart; \textbf{reduce condition}, which determines when to reduce; \textbf{rephase function}, which manages rephase heuristics; \textbf{rephase condition}, which determines when to rephase; \textbf{bump var activity}, which governs the order of variables selection in Make Decision; and \textbf{bump cla activity} that govern the order of clauses being removed during reduce. 
These functions collectively improve the solver's ability to handle large and complex SAT instances by balancing exploration, exploitation, and resource management.  The combination of the conflict-driven learning, backtracking, and heuristic enhancements makes ModSAT a robust SAT solver.

\subsection* {Three principles behind ModSAT}
As noted earlier, the high complexity of modern SAT solvers makes them difficult for LLMs to optimize. For instance, modifying a single variable may require changes across multiple locations, hindering the generation of correct code. Additionally, to enable localized heuristic modifications without disrupting unrelated components, adequate error tolerance must be allowed, which requires comprehensive variable context sharing for the modularized solver. To address these challenges, we have formulated three guiding principles:
\begin{itemize}
    \item maintain functions simple and focus;
    \item use class variables for shared information;
    \item proactively prevent bugs during heuristics discovery.
\end{itemize}

\textbf{Maintain functions simple and focus} means that the function optimized by LLMs should be simple and explicit, which is uncommon in complex solvers. Contemporary solvers, typically implemented in C/C++, often employ complex data structures coupled with poorly modularized functions. This structural deficiency can mislead LLMs on variable scopes, so  erroneous codes may be generated when optimizing the solvers. 

\textbf{Use class variables for shared information} means that class member variables are more friendly than global variables for LLMs to access. Although functions typically utilize only a few variables in the original solver implementations, LLMs tend to coordinate with additional variables to strengthen the diversity of heuristics. Therefore, instead of providing the information on all possible variables in the prompt to {LLMs},  only relevant class member variables are provided as contextual information, which allows them to autonomously determine which variables to utilize.

\textbf{Proactively prevent bugs during heuristics discovery} means that we should fix the bugs written by LLMs proactively, so that the solver can compile correctly with the same heuristics. Many compilation errors admit straightforward resolutions, particularly those arising from common issues such as missing inclusions or variable scope misinterpretations. Fixing the bug proactively is helpful for LLMs to generate more diverse and correct codes.

\subsection*{Presearch strategy}
The goal of the presearch strategy is to prune low-value function candidates before the heuristic search.
Previous work~\cite{autosat} usually utilizes strategies such as evolutionary algorithms (EA) and greedy hill climber to optimize all the candidate functions, which becomes difficult to scale as the number of functions  grows. For instance, in optimizing heuristic functions as a pseudo-Boolean optimization problem (e.g., LLMs provide merely two candidates for each function) with sequential dependency, EA requires approximately $0.54n^2$ ($n$ is the number of functions need to be optimized) trials of generating new heuristics functions as illustrated in recent empirical studies\cite{doerr2018towards, ye2020benchmarking}. However, the number of candidates for each function that LLMs can provide is more than two in practice, and the performance contribution of each function may depend on complex dependencies. These facts indicate that significant optimization time is required for achieving the optimal results~\cite{doerr2018static, doerr2019theory}. Since evaluating a SAT solver is a time-consuming task with a timeout bound of $5,000$ seconds, searching for the optimal combination of SAT heuristic functions is computationally prohibitive.  

To address this issue, we  propose a presearch strategy to mitigate the combinatorial explosion:  prune function candidates through small-scale preliminary tests (single-function evaluation on compact datasets), and then execute $(1+\lambda)$ EA (with $\lambda=1$) search on the refined subset of functions. Our empirical analysis shows that, for a given dataset, some functions consistently degrade solver performance, which suggests a natural dimensionality-reduction strategy to accelerate the search for effective heuristics. This two-phase approach significantly reduces the iteration counts while preserving optimization effectiveness. Note that the $(1+\lambda)$ EA~\cite{doerr2019theory} is an evolutionary algorithm where, in each generation, a single parent individual produces $\lambda$ offspring through mutation, and the fittest individual among the parent and all offspring is selected as the parent for the next generation, ensuring elitist selection.

More precisely, we construct a representative subset comprising $50\%$ of the problem instances from the original dataset. On this compact subset, each candidate function's standalone impact based on the PAR-2 metric (where lower values indicate better solver performance) is evaluated. Typically, this phase retains about four high-impact functions that consistently improve the PAR-2 score, and functions that degrade the PAR-2 score are identified and pruned. For each dataset, the pruned functions are not considered in the subsequent evolutionary search. In practice, this phase typically retains a small set of high-impact functions (e.g., four functions) that consistently show a positive or neutral effect on PAR-2 in these preliminary tests. Subsequently, we execute a $(1+\lambda)$ EA  on the full dataset using this refined function set in the following heuristics discovery process. The convergence behavior of AutoModSAT with and without the presearch (see Supplementary Figure 8) show that the presearch can improve the computational efficiency of AutoModSAT.

\begin{figure}[ht!]
    \centering
        \includegraphics[width=\textwidth]{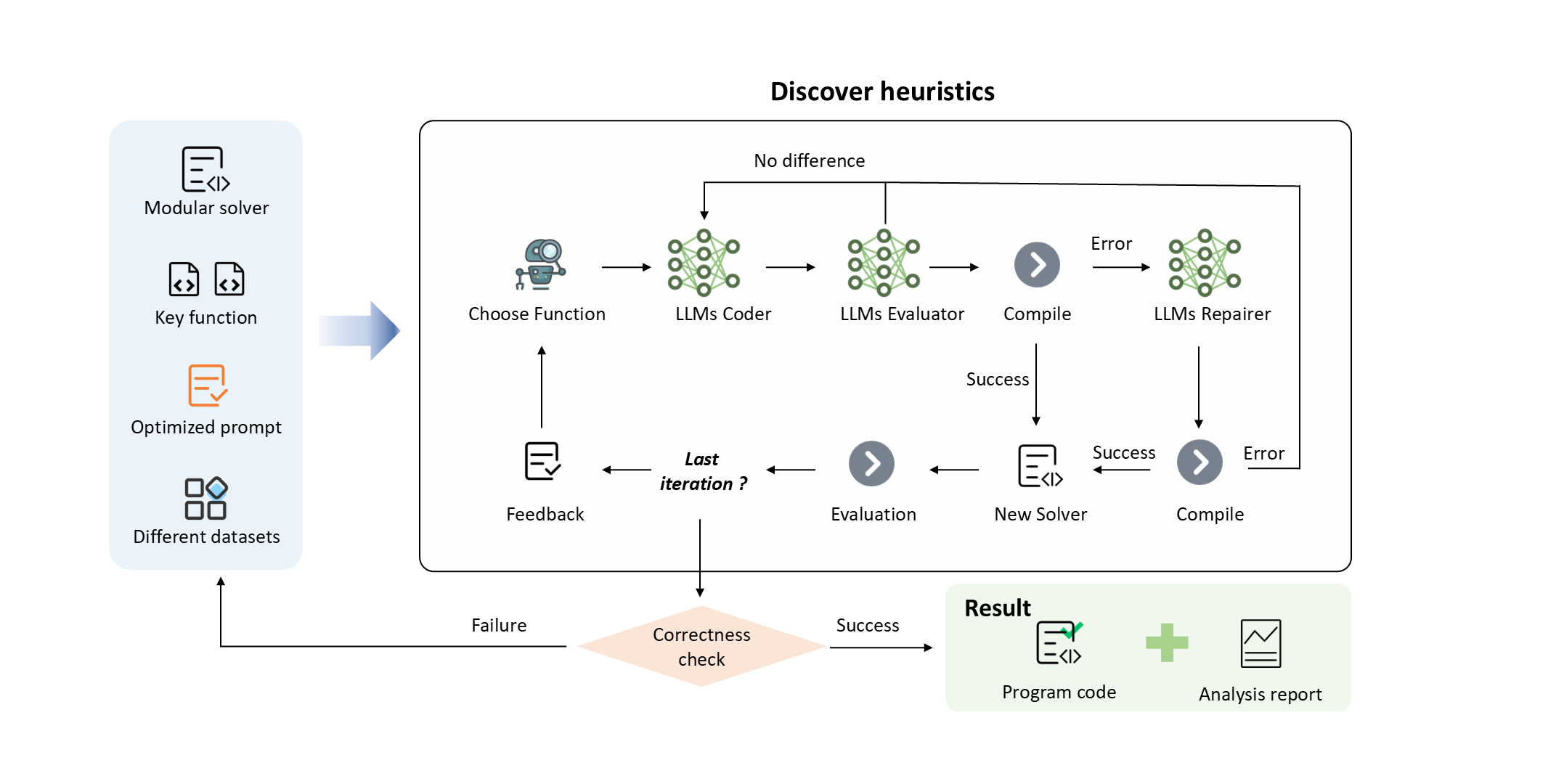}
\caption{\textbf{Heuristic discovery.} This procedure iteratively implements LLMs agents to generate new heuristics. Heuristics that passed a synonymous check are evaluated on the datasets and higher-performing heuristics are integrated back into the new solver, while failures trigger automatic repair. The loop continues until the iteration budget is exhausted.}
\label{fig:discovery}
\end{figure}

\subsection*{Prompt template and  optimization}
For clarity, we first introduce the basic prompt template, then explain the motivation for prompt optimization, and finally describe the optimization procedure. We begin with the basic prompt template for the LLM agents, which follows the instructions in OpenAI framework docs~\cite{openai_docs} and has the following format:
\begin{itemize}
    \item Define the \textbf{Role} of an agent as a solver expert who needs to assess and improve the heuristics function in the SAT solver.
    \item Clearly state the \textbf{Goal}, such as providing optimization suggestions, writing code, or feedback.
    \item Enhance the agents' capabilities by inserting optional \textbf{Tips} that guide them to avoid common mistakes during code generation. Additionally, through this flexible interface, agents can effectively utilize external codes and results and can be instructed to specify the types of modification directions such as changing parameters, modifying heuristics, or adding new heuristics.
    \item Key code of SAT solver is appended at the end of each prompt to ensure all agents are in the same context. Note that the key code includes member variables in the cpp class of the solver (LLMs may utilize), along with the main loop function, and all corresponding functions LLMs may need to understand (e.g., such as the functions which would be called by the target optimized function).
\end{itemize}

The basic prompt template only provides a unified interface for different agents, yet identifying effective prompts remains challenging in practice~\cite{promptengineering}. Traditional approaches require extensive trial-and-error by experts to identify effective prompts, a process further complicated by the dynamic nature of optimal prompts as LLMs evolve. While conventional prompt optimization leverages supervised learning when labels can be obtained easily~\cite{ape, zhou2022large, pryzant2023automatic}, our task faces two critical constraints: prohibitive execution time ($5000$ seconds timeout per instance), and the absence of definitive labels due to the performance variance across different datasets.

To address these issues, an unsupervised method that utilizes Shannon entropy as an evaluation metric is adopted for automatic prompt optimization here. This metric quantifies the population diversity at each time step based on clusters of all the individuals. Firstly, we employ the CodeT5+ embedding model~\cite{codet5+} to generate embeddings for the synthesized code, which is preprocessed according to the Google C++ style guide~\cite{googlestyle} (e.g., removing comments and ensuring uniform indentation). The resulting embeddings form the set $\mathbf{X}$, where each code sample $x \in \mathbf{X}$ is represented by an $m$-dimensional vector. 
Then, we use an unsupervised learning method to cluster the $N$ distinct code embeddings generated by the same prompt. Specifically, we apply the K-Means++~\cite{kmeans++} to partition the embeddings into $K$ clusters. More precisely, given the embeddings $\mathbf{X} = \{x_1, \dots, x_N\} \subset \mathbb{R}^m$, K-Means++ consists of the following three steps:
\begin{enumerate}
    \item[(1)] 
Initialize centroids: Select the first centroid $c_1$ uniformly at random from $\mathbf{X}$. For each subsequent centroid $c_j$ ($j=2,\dots,K$), choose it as $x \in \mathbf{X}$ with probability proportional to $\min_{k=1}^{j-1} |x - c_k|^2$;
\item[(2)] Assign clusters: Assign each embedding $x_i$ to the nearest centroid $c_j$ using Euclidean distance.
\item [(3)]Update centroids: Recompute the centroids as the mean of all embeddings in each cluster, and then repeat step (2).
\end{enumerate}

After the clustering is completed, we calculate the probability distribution of each cluster, denoted $p_i = \frac{|C_i|}{N}$ ($i=1,\cdots K$), where \( |C_i| \) is the number of individuals of the $i$-th cluster. Then the entropy, \( H(X) \) is given by $H(X) = -\sum_{i=1}^{K} p_i \log(p_i)$.
This criterion is able to eliminate the dependency on explicit labels while maintaining adaptability to the evolution of LLMs, particularly suitable for compute-intensive optimization tasks with ambiguous success criteria.

Based on the Shannon entropy metric, the three components in the prompt templates can be  optimized by LLMs. In each iteration, we randomly select a component to optimize by LLMs, compute the correctness and diversity of the generated codes, and update the component in prompt with better performance in both correctness and diversity. This loop will iterate until the stopping criteria is met, see Supplementary Algorithm~3 for details.

\subsection*{Heuristics discovery}
After obtaining the optimized prompt, the modularized SAT solver, and the presearch function candidates, AutoModSAT starts to discover heuristics in different datasets. This phase involves three LLM-based agents, including LLMs coder to generate heuristics, LLMs evaluator to filter out semantically equivalent code and avoid redundant search iterations, and LLMs repairer to correct common errors from LLM-generated code. See Figure~\ref{fig:discovery} for an illustration of this process. 

Within each iteration of this process, a specific function is selected from the effective candidate space. The LLMs coder, whose prompt has been optimized in the automatic prompt optimization stage, is then called to generate code. Since we have observed that the LLMs coder may generate code that is syntactically different but functionally synonymous to the original implementation, the generated code is then evaluated by the LLMs evaluator. The code that passes the synonymous evaluation will be executed immediately against a predefined dataset. Successful execution triggers a performance evaluation phase to assess the heuristic's efficacy.

If the code execution fails, the code along with the error output are sent back to LLMs repairer. LLMs repairer utilizes this feedback to generate revised code and attempt to repair the bug. However, if the repair attempt fails, then the current iteration for the selected function candidate is terminated and the process restarts with a new iteration, possibly selecting a different function candidate to optimize. Moreover, when the code execution succeeds, it will be evaluated and compared with existing heuristics. If the new heuristic demonstrates a superior performance, it is dynamically integrated into the modularized solver to enhance its capabilities.  

Afterwards, we  additionally verify the correctness  of the discovered solver. For SAT instances, the solver outputs a satisfying assignment, which is checked against the original CNF formula. For UNSAT instances, the solver outputs a DRAT UNSAT proof that is verified using drat-trim~\cite{drat-trim}. Any solver candidate that fails this correctness check would be discarded. More details about the correctness verification can be found in Supplementary Section 4.7. 

In summary, the entire process consists of candidate selection, LLM-driven code generation, execution validation, bug repair, performance evaluation, and solver update, followed by the correctness  verification. This structured framework facilitates systematic and adaptive code generation, and can successively refine the solver's heuristic search space through successive iterations.

\section*{Data Availability}
The datasets used in this study are available from the following sources. The benchmark instances from the SAT Competition 2023 and 2024 are publicly available at the official competition website: \url{https://satcompetition.github.io/}. The datasets generated by the Picat solver are available in the Zenodo repository: \url{https://doi.org/10.5281/zenodo.17856423}. The proprietary dataset from real industrial EDA scenarios is also included in the same Zenodo repository: \url{https://doi.org/10.5281/zenodo.17856423}.

\section*{Code Availability}
The source code for AutoModSAT has been deposited in the Zenodo repository under the MIT License and is publicly available at: \url{https://doi.org/10.5281/zenodo.17856395}.

\bibliography{main}

\section*{Acknowledgments}
We thank the anonymous reviewers for their constructive comments, which helped improve the manuscript. We also thank our colleagues for helpful discussions on SAT solver optimization and experimental evaluation.

\section*{Funding}
This work was supported by the National Natural Science Foundation of China (NSFC) Key Program under Grant No. E6JZ500101.

\section*{Author contributions}
Y.S. performed the main experiments, wrote the manuscript, and coordinated the overall progress of the project. F.Y. contributed to manuscript writing and editing. Z.C. contributed to some of the experimental work. K.W. contributed to manuscript writing and provided conceptual guidance. S.C. contributed to manuscript editing and provided conceptual guidance. K.W. and S.C. jointly supervised the project.

\section*{Competing interests}
The authors declare no competing interests.

\newpage

\begin{center}
{\Large\textbf{Supplementary Materials}}

\vspace{1cm}

{\Large{Discovering heuristics in a complex SAT solver with large language models}}
\vspace{1cm}

Yiwen Sun, Furong Ye, Zhihan Chen, Ke Wei, Shaowei Cai
\vspace{1cm}

\end{center}
\section{{Additional Background on SAT}}

Let $V = \{x_1, x_2, \dots, x_n\}$ be a set of Boolean variables. Recall that  \textit{literal} is either a variable $x$ or its negation $\neg x$, a \textit{clause} is a disjunction of literals, while \textit{conjunctive normal form} (CNF) formula $F = C_1 \land C_2 \land \dots \land C_m$ is a conjunction of clauses. For simplicity we assume all clauses are non-tautological. In other words, no variable $x$ occurs positively $(x \in C)$ and negatively $(\neg x \in C)$ in the same clause.

A (partial) mapping $\alpha : V \to \{0, 1\}$ is called an \textit{assignment}. If $\alpha$ maps all variables to a boolean value, it is termed a \textit{complete} assignment; otherwise, it is referred to as a \textit{partial} assignment. The value of a variable $x$ under an assignment $\alpha$ is denoted as $\alpha[x]$. An assignment $\alpha$ satisfies a clause if at least one literal evaluates to true under $\alpha$, and satisfies a CNF formula if it satisfies all its clauses. A CNF formula $F$ is satisfiable if there is at least one satisfying assignment. The empty clause $\bot$ is always unsatisfiable and represents a \textit{conflict}. SAT is the problem of deciding whether a given CNF formula is satisfiable.

To solve  the SAT problem, a straightforward approach that is implemented in a complete SAT solver is to conduct exhaustively search over all possible truth assignments to the variables in $V$ to determine if at least one satisfies $F$ (proving $F$ is satisfiable). If no satisfying assignment exists, the solver conclusively proves $F$ is unsatisfiable. Within this framework, the strategies for branching (for choosing which variable to assign next and what value to assign it) are crucial. Key techniques to enhance the search efficiency include restart strategies, restart strategies, and activity-based heuristics (often implemented via the bumping variable or clause activity scores).

\subsection{CDCL solver}

Conflict-Driven Clause Learning (CDCL)~\cite{cdcl} is the most common approach and plays a dominant role in modern high-performance SAT solvers, see Supplementary Algorithm~\ref{alg:CDCL} for a basic CDCL framework. The core idea is conflict analysis: when a conflict occurs, the solver analyzes it and derives a new clause that explains the inconsistency. These derived clauses, called learned clauses, enable the solver to prune large portions of the search space, which significantly improve the computational efficiency. However, an excessive number of learnt clauses can degrade unit propagation speed and thus consume excessive memory. Consequently, identifying high-quality learned clauses and reducing their quantity are essential for maintaining a solver's performance.


Specifically, CDCL solvers operate on a propagation and learning mechanism, complemented by decision heuristics. The implementation of propagation and learning is a standard practice and overall similar in different solvers. In contrast, the decision policies (for variable selection) and restart policies (for search restarts) which are crucial for performance vary significantly. Due to the need for manual design and the absence of rigorous mathematical proofs for their optimality, these policies are referred to as important heuristics.

There is a long history of research on these heuristics in SAT solvers, among which branching heuristics play a crucial role and continue to impact the performance of CDCL solvers. For example, Variable State Independent Decaying Sum (VSIDS)~\cite{vsids} is a family of branching heuristics that seek to assign a value to the most promising variable in the Make Decision phase. Another important branching heuristic is Learning-Rate Branching (LRB)~\cite{lrb}, which frames branching as an optimization problem that picks a variable to maximize a metric called learning rate.

Restart heuristics are also essential for enhancing the performance of CDCL solvers. It allows the solver to abandon the current search path and backtrack to a specific decision level. In this process, the learnt clauses are usually maintained for the next search. Fast restart~\cite{ramos2011between} is a widely used method, and Luby restarts~\cite{luby} is also heavily used because they represent a prior optimal strategy. However, most implementations are now switched to Glucose-style restarts~\cite{glucose}, which are widely used in the SAT Competition. For more details, see the extensive overview given by Armin Biere~\cite{biere2015evaluating}.

Rephase is another technique in CDCL solvers. The primary objective is to reset or adjust the current partial assignment, thereby enabling the solver to explore a diverse search space. PrecoSAT and PicoSAT~\cite{precosat} utilize a Jeroslow-Wang score~\cite{jeroslow1990solving}, computed from either all clauses or only irredundant clauses, with rephasing intervals scheduled according to a Luby sequence. StrangeNight~\cite{Strangenight} employs a strategy of flipping values with a certain probability that depends on the depth of the assignment. The motivation is to avoid the heavy-tail phenomenon.  These rephasing heuristics have been used and compared in the SAT solver Riss~\cite{balint2015overview}.

\begin{algorithm2e}[t]
\caption{CDCL Framework}
\label{alg:CDCL}
\textbf{Input:} A CNF $F$ of SAT instance\;
\textbf{Initialization: } decision level $d \leftarrow 0$, current assignment of variables $\mathcal{X} \leftarrow \emptyset$\;

\While{\textbf{True}}{
$\mathcal{X} \leftarrow \textbf{Unit Propagation}(F,\mathcal{X})$\;
\eIf{Conflicts are detected in $\mathcal{X}$}{
\eIf{$d == 0$}{return \textsc{UNSAT}}
{$\mathcal{C}_{conflict}, d_{backtrack} \leftarrow \textbf{Analyze Conflict}(F,\mathcal{X})$ \;
$C_{learned} \leftarrow \textbf{Learn Clause}(\mathcal{C}_{conflict},\mathcal{X})$ \;
$F \leftarrow F \land C_{learned}$\;
$\mathcal{X} \leftarrow \textbf{Backtrack}(\mathcal{X},d_{backtrack})$\;
$d \leftarrow d_{backtrack}$\;
}
}
{
\eIf{All variables are assigned with a value}{return \textsc{SAT}}
{
$d \leftarrow d + 1$\;
$\mathcal{X} \leftarrow \textbf{Make Decision}(\mathcal{X})$
}
}
}
\end{algorithm2e}

\section {Illustration of Three Principles behind ModSAT}
As already mentioned in the paper,  there are three principles when developing ModSAT to ensure it is LLM-friendly:

\begin{itemize}
    \item \textbf{Maintain functions simple and focus.} The function optimized by LLMs should be simple and explicit, unlike common implementations in complex solvers.
    \item \textbf{Utilize class variables for shared information.} Local variables should be declared as class member variables to give LLMs access to them.
    \item \textbf{Proactively prevent bugs during heuristics discovery.} The bugs written by LLMs should be fixed proactively, so that the solver could compile correctly with the same heuristics, which helps LLMs to generate more diverse correct codes.
\end{itemize}

Supplementary Figure~\ref{fig:rule1-2} demonstrates the application of the principle \emph{Maintain functions simple and focus},  with the original function given in Supplementary Figure~\ref{fig:rule1-1}. Instead of modifying the whole search function by LLMs, we modularize it into three distinct functions by isolating the components  that significantly impact the performance and are suitable for LLMs to modify. The decomposition also enables LLMs to focus on refining one kind of heuristic at a time, thereby enhancing the code generation capability.



\begin{figure}[ht!]
\begin{cppcode}{Maintain functions simple: Original function to modify}
lbool Solver::search(int nof_conflicts){
    // if there is a conflict
    ...... ......
    // if there is no conflict
    if ((lbd_queue_size == 50 && 0.8 * fast_lbd_sum / lbd_queue_size > slow_lbd_sum / conflicts) || !withinBudget())
        restart_function();

    // Simplify the set of problem clauses:
    if (decisionLevel() == 0 && !simplify())
        return l_False;
            
    // Reduce the set of learnt clauses:
        if (learnts.size()-nAssigns() >= max_learnts)
                reduceDB();
    if (rephase_condition())
        rephase_function();
        
    Lit next = lit_Undef;
    while (decisionLevel() < assumptions.size()){
        // Perform user provided assumption:
        Lit p = assumptions[decisionLevel()];
        if (value(p) == l_True){
            // Dummy decision level:
            newDecisionLevel();
            }else if (value(p) == l_False){
                analyzeFinal(~p, conflict);
                return l_False;
            }else{
                next = p;
                break;
            }
        }

    if (next == lit_Undef){
        // New variable decision:
        decisions++;
        next = pickBranchLit();

        if (next == lit_Undef)
            // Model found:
            return l_True;
    }
    newDecisionLevel();
    uncheckedEnqueue(next);
}
\end{cppcode}
    \caption{Illustration of principle {\em maintain functions simple and focus.}}
    \label{fig:rule1-1}
\end{figure}

\begin{figure}[ht!]
\begin{cppcode}{Maintain functions simple: modularized function to modify}
// original function which has been modularized
lbool Solver::search(int nof_conflicts){
    ...... ......

    if (restart_condition())
        restart_function();
    if (reduce_condition())
        reduceDB();

    if (rephase_condition())
        rephase_function();
        
    ...... ......
}

// functions to modify
bool Solver::rephase_condition() {
    if (rephases >= rephase_limit) 
        return true;
    else 
        return false;
}

bool Solver::reduce_condition() {
    if (rephases >= rephase_limit) 
        return true;
    else 
        return false;
}

bool Solver::restart_condition(){
    if ((lbd_queue_size == 50 && 0.8 * fast_lbd_sum / lbd_queue_size > slow_lbd_sum / conflicts) || !withinBudget())
        return true;
    else
        return false;
}
\end{cppcode}
    \caption{Illustration of principle {\em maintain functions simple and focus}.}
    \label{fig:rule1-2}
\end{figure}

For the principle \emph{utilize class variables for shared information}, we move variables that may reside in local scope into class members to ensure that LLMs can access this shared information, see Supplementary Figure~\ref{fig:rule2-1}.

\begin{figure}[ht!]
\begin{cppcode}{Add class member variables}
    // LBD heuristics
    int lbd_queue[500],   // circled queue saved the recent 500 LBDs.
        lbd_queue_size,   // The number of LBDs in this queue
        lbd_queue_pos;  
    double fast_lbd_sum, slow_lbd_sum;   

    // rephase heuristics
    int rephases, rephase_limit, rephase_count, threshold;
    double last_rephase_progress;

    // restart heuristics
    int curr_restarts;
    double last_restart_progress;
\end{cppcode}
    \caption{Illustration of principle {\em utilize class variables for shared information.}}
    \label{fig:rule2-1}
\end{figure}

Supplementary Figures~\ref{fig:rule3-1} and Supplementary \ref{fig:rule3-2} provide two examples of how to {\em proactively prevent bugs during heuristics discovery}. The first one is to include some extra packages to prevent LLMs from implementing extra functions, while the second one is to overload common functions to prevent LLMs from incorrectly implementing simple functions due to a misunderstanding of data structures.

\begin{figure}[ht!]
\begin{cppcode}{Prevent bugs: add more packages}
#include <math.h>
#include <unordered_set>
#include <algorithm>
using namespace std;
\end{cppcode}
    \caption{Illustration of principle {\em proactively prevent bugs during heuristics discovery.}}
    \label{fig:rule3-1}
\end{figure}

\begin{figure}[ht!]
\begin{cppcode}{Prevent Bugs: overloading functions}
#include <type_traits>

template <typename T1, typename T2>
typename std::common_type<T1, T2>::type max(T1 a, T2 b) {
    static_assert(std::is_integral<T1>::value && std::is_integral<T2>::value,
                  "max: Both types must be integers (int or long int)");
    return (a < b) ? b : a;
}

template <typename T1, typename T2>
typename std::common_type<T1, T2>::type min(T1 a, T2 b) {
    static_assert(std::is_integral<T1>::value && std::is_integral<T2>::value,
                  "min: Both types must be integers (int or long int)");
    return (a < b) ? a : b;
}

template <typename T1, typename T2>
typename std::common_type<T1, T2>::type max(T1 a, T2 b) {
    static_assert(std::is_floating_point<T1>::value && std::is_floating_point<T2>::value,
                  "max: Both types must be floating-point (float or double)");
    return (a < b) ? b : a;
}

template <typename T1, typename T2>
typename std::common_type<T1, T2>::type min(T1 a, T2 b) {
    static_assert(std::is_floating_point<T1>::value && std::is_floating_point<T2>::value,
                  "min: Both types must be floating-point (float or double)");
    return (a < b) ? a : b;
}
\end{cppcode}
    \caption{Illustration of principle {\em proactively prevent bugs during heuristics discovery.}}
    \label{fig:rule3-2}
\end{figure}

\section{More on   Presearch Strategy and Automatic Prompt Optimization}

The detailed description of the presearch strategy can be found in Supplementary Algorithm~\ref{alg:autosat-pre} and Supplementary Algorithm~\ref{alg:autosat-ea}. In addition, we also investigate the contribution of each function to the final results by calculating the number of times that a function contributes to the final performance improvement in each experiment, see Supplementary Figure~\ref{fig:functioncontribution}, where each subfigure shows the results for one dataset. It can be observed that almost all the selected functions  contribute substantially to the final performance.

The pseudocode for automatic prompt optimization is presented in Supplementary Algorithm~\ref{alg:prompt}. Examples of the original prompt and the one after optimization are provided in Supplementary Figure~\ref{fig:coder_prompt}.

\begin{algorithm2e}[ht!]
\caption{Presearch Strategy in AutoModSAT}\label{alg:autosat-pre}
\textbf{Input:} Datasets $P$, modularized SAT solver with seven functions $\{h_1,\ldots,h_7\}$, prompt template, baseline functions $\{b_1,\ldots,b_7\}$ \\
$P_{\text{compact}} \leftarrow$ 50\% representative instances from $P$ \\
$R \leftarrow \emptyset$  \tcp*{Retained function set}
\For{each function $h_i \in \{h_1,\ldots,h_7\}$}{
    $A_{\text{test}} \leftarrow$ build solver replacing $h_i$ with baseline $b_i$ \\
    $f_i, s_i \leftarrow$ evaluate($A_{\text{test}}$, $P_{\text{compact}}$) \tcp*{Get PAR-2 metric} 
}
$F \leftarrow $ sort($f_1, ..., f_7$) \tcp*{Sort PAR-2 metric in different functions}
Get top 4 function index $R$ from $F$ \\

\end{algorithm2e}

\begin{algorithm2e}[ht!]
\caption{Heuristics Discovery Strategy in AutoModSAT}\label{alg:autosat-ea}
\textbf{Input:} Datasets $P$, modularized SAT solver with pruned four functions $\{h_1,\ldots,h_4\}$, prompt template, baseline functions $\{b_1,\ldots,b_4\}$ \\\
$A \leftarrow$ solver with functions: $(\forall i \in R: h_i) \cup (\forall i \notin R: b_i)$ \\
$f^* \leftarrow$ evaluate($A$, $P$)  \tcp*{Full dataset evaluation}

$\text{evalBudget} \leftarrow 50$  \tcp*{Maximum evaluations}

\While{evalBudget $>0$}{
    $M \leftarrow \emptyset$ \\
    $\ell \sim \text{Bin}(|R|,\frac{1}{|R|})$  \tcp*{Sample modification count based on (1+$\lambda$)~EA}
    chosen $\ell$ distinct values $M \leftarrow \{m_0, \ldots, m_\ell\}$ from $R$ uniformly at random\;
    Generate new functions $\{h'_m\}_{m\in M}$ via LLMs using $A$ and $M$ \\
    $A' \leftarrow$ update $A$ with $\{h'_m\}_{m\in M}$ \\
    $f(A') \leftarrow$ evaluate($A'$, $P$) \\
    \If{$f(A') \leq f^*$}{
        $A \leftarrow A'$ \\
        $f^* \leftarrow f(A')$
    }
    $\text{evalBudget} \leftarrow \text{evalBudget} - 1$
}
\end{algorithm2e}

\begin{algorithm2e}[t]
\caption{Automatic Prompt Optimization}
\label{alg:prompt}
\SetAlgoLined
\DontPrintSemicolon
\KwIn{initial prompt template $P$, solver codebase $S$, max\_iterations $i$, prompt optimized part $R =\{ Role, Goal, Tips \}$}
\KwOut{optimized prompt template $P^*$}
$i \gets 10$ \tcp*{prompt optimization iterations}
$j \gets 20$ \tcp*{number of code generation in each iteration}
\While{$i \ge 0$}{
    select prompt part $r$ from $R$ uniformly at random \\
     $P' \gets$ refine\_prompt($r, P$) \tcp*{LLMs refine the prompt template}
    \While{$j \ge 0$}{
    generated code $c \gets \text{call\_llm}(\textit{current\_prompt})$ \\
    compilation error $e \gets$ compile($c$, $S$) \\
    \eIf{$\neg e $}{
            $C \gets C \cup c$
            }
    {
        
        $\text{execute\_code}(\textit{corrected\_code})$ \tcp*{execute code}\;  
    }
    $j \gets j - 1$\;
    }
            
    diversity $ d_i \gets \text{compute\_code\_diversity}(C)$\;
    success rate $ s_i \gets \text{compute\_code\_success}(C)$\;

    \eIf{$d_i > d$ and $s_i > threshold $}{
        $d \gets d_i $\\
    $P \gets \text{update\_prompt}(S, P')$\;
    }

    $i \gets i - 1$\;
}
\end{algorithm2e}

\begin{figure*}[t]
    \centering
        \includegraphics[width=\textwidth]{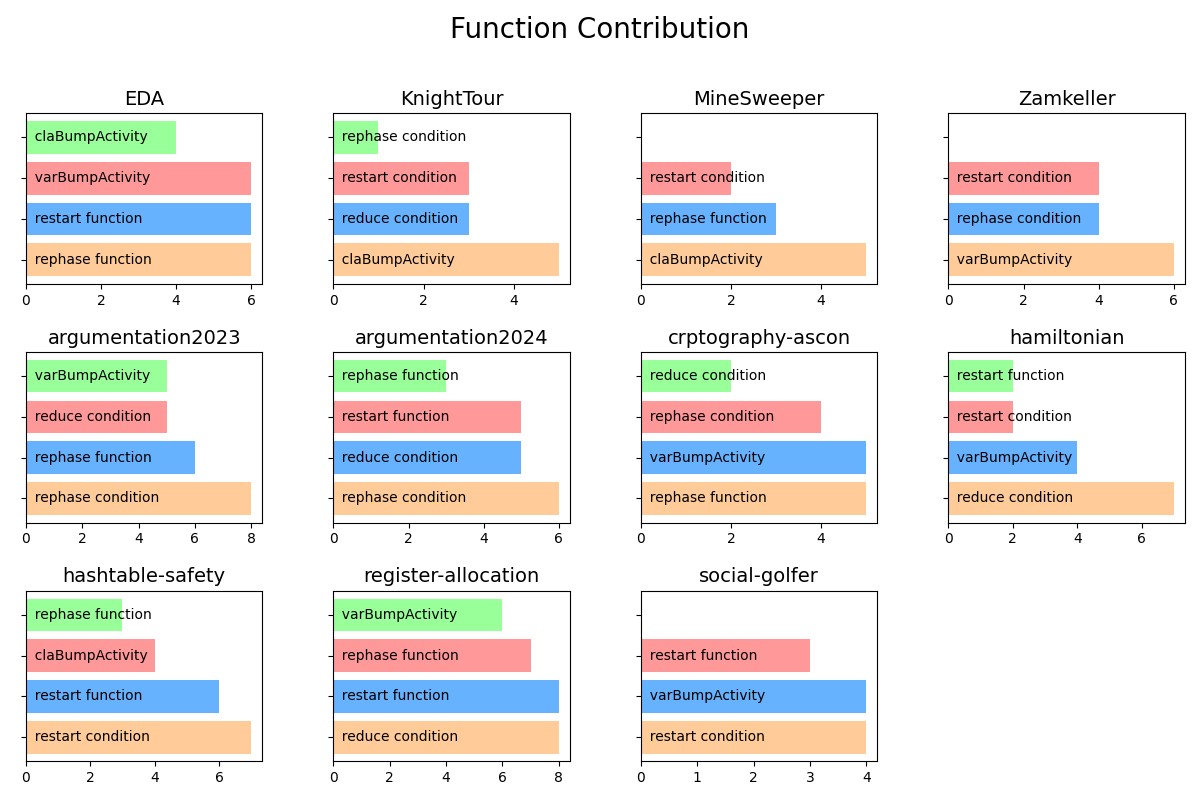}
\caption{Function contribution: where the vertical axis lists different function candidates  and the horizontal axis denotes the frequency of contributions.}

\label{fig:functioncontribution}
\end{figure*}

\begin{figure}[ht!]
\begin{tcolorbox}[colback=gray!10!white, colframe=gray!50!black, title=Original Prompt]
(Role) You are a SAT solver researcher trying to rewrite the \{\{ func\_name \}\}  function(s). \\

(Goal) Your goal is to improve the SAT solver by rewriting the \{\{ func\_name \}\}  function(s), after reading and understanding the <key code> of SAT solver below.\\

(Tips) Tips:\\
1) Your rewritten function code must start with '''// start {function name}''' and end with '''// end {function name}''' \\
2) Your rewritten function(s) code must be different from original code, not just rewrite code synonymous! \\
3) You are not allowed to create your own new function(s) in the rewritten function(s).  You are not allowed to create your own new global variables, but you can use the global variables existing in the <key code>. \\
4) Make sure the rewritten function(s) code can be executed correctly. \\

<key code> of SAT solver is:\\
\{\{ replace\_key\_code \}\}

\end{tcolorbox}

\begin{tcolorbox}[colback=gray!10!white, colframe=gray!50!black, title=Updated Prompt]
(Role) You are a SAT solver researcher trying to improve the \{\{ func\_name \}\} function. \\

(Goal) Objective:\\
Your goal is to improve the SAT solver by rewriting the \{\{ func\_name \}\}  function. \\

Instructions: \\
1. Carefully read and comprehend the <key code> of the SAT solver provided below. \\
2. Analyze potential improvements and devise a strategy for optimizing the heuristics of function. \\
3. Deliver your improved function(s) with the following format:\\
   - Begin with: `// start {function name}`\\
   - End with: `// end {function name}`\\

(Tips) Tips:\\
1. Ensure that your rewritten function(s) are substantially different from the original, beyond mere synonym replacements.\\
2. You may utilize existing global variables from the <key code>, but refrain from introducing new global variables.\\
3. Verify that the rewritten function(s) execute correctly.\\

Take a deep breath and think it step by step. \\

<key code> of SAT solver is: \\
"""
\{\{ replace\_key\_code \}\}
"""
...
\end{tcolorbox}
    \caption{Comparison of original prompt and optimized prompt}
    \label{fig:coder_prompt}
\end{figure}


\section{Experimental details}

\subsection{Dataset description}
\label{sec:dataset}
Here we give more details about the datasets, including 7 datasets selected from SAT Competition 2023 and 2024, 3 generated ones by Picat, and another one from an industrial EDA scenario. 
\begin{itemize}
    \item \textbf{Argumentation} involves finding acceptable sets of arguments in a directed graph where attacks between arguments are represented by edges.
    
    \item  \textbf{Social Golfer} is a combinatorial problem that aims to assign golfers to groups over several weeks, ensuring no two golfers play in the same group more than once.
    
    \item  \textbf{Hashtable Safety} focuses on verifying the correctness of operations in a hash table to avoid collisions and ensure the integrity of the structure.

    \item \textbf{Register Allocation} is a problem that arises in compiler optimization, where the goal is to assign a limited number of CPU registers to variables in a program. 
    
    \item \textbf{Cryptography-Ascon} is a lightweight cryptographic algorithm challenge which focuses on implementing and verifying the Ascon cipher, a NIST-standardized authenticated encryption scheme, for resource-constrained IoT devices, balancing security against differential attacks with minimal computational overhead.
    
    \item \textbf{Hamiltonian} is a graph theory problem that involves determining whether a given graph contains a Hamiltonian cycle, a closed loop visiting each vertex exactly once, which is NP-complete and often applied to route optimization or circuit design verification.

    \item \textbf{MineSweeper} is derived from the classic MineSweeper game, where the objective is to determine the placement of hidden mines on a grid based on numerical clues. 
    
    \item \textbf{LangFord} is a combinatorial  problem that is about finding a specific permutation of the sequence $1, 1, 2, 2, ..., n, n$ where the two copies of each number $k$ are exactly $k$ units apart.

    \item \textbf{KnightTour} aims to find a path for a knight on a chessboard that visits every square exactly once, with possible extensions to different board sizes and types.
    
    \item \textbf{Zamkeller} involves finding a permutation of integers from $1$ to $n$ that maximizes the number of differential alternations in subsequences divisible by integers from $1$ to $k$, where $1 < k < n$.

    \item \textbf{EDA} involves formally proving whether two design specifications are functionally equivalent, which is one of the most essential techniques in Electronic Design Automation and digital IC design. It has a wide range of applications, such as functional equivalent logic removal, sequential equivalence checking, circuit-based method for symmetries detection, engineering change orders, among others.

\end{itemize}

For the generated instances using Picat~\cite{picat}, we adopt  the settings in Chapters 2 and 3 of the book by Nengfa Zhou~\cite{picat} and conduct grid sampling within a parameter space. Specifically, the parameter space  $\Theta = \{\theta_1, \theta_2, \ldots\}, \theta^{L}_i \le \theta_i \le \theta^{U}_i$ is defined to ensure that three baseline solvers including MiniSat, CaDiCal and Kissat, can obtain a solution within proper time range, i.e., [$1$, $5000$] seconds. Then we apply a space $\Theta'$ to generate the dataset by enlarging the upper bound of $\Theta$, e.g., $\theta'^U_i = \theta'^U_i * 1.2$, such that we can test whether AutoModSAT can solve the instances where the baseline solvers cannot.

\begin{itemize}
    \item \textbf{MineSweeper} \\ 
    Parameter $\Theta$: $\{m, n, k, p\}$;	\\
    Parameter Space: $[500, 1600] \times [400, 3200] \times [72689, 1572118] \times [0.32, 0.38]$; \\	
    Notes: $m, n$ represent the grid size of the Minesweeper game, $k$ is the total number of mines, and $p$ is the probability that a given cell contains a mine (range: $0.32$ to $0.38$).

    \item \textbf{KnightTour} \\
    Parameter $\Theta$: $\{k\}$; \\
    Parameter Space: $[12, 75]$; \\
    Notes: A $k \times k$ chessboard where a knight's tour is attempted, covering all squares and returning to the start point (a solution is not possible for odd-sized boards, i.e., they are unsatisfiable).
    \item \textbf{Zamkeller}: \\
    Parameter $\Theta$: $\{k, n\}$;\\
    Parameter Space: $[3, 34] \times [25, 100]$; \\
    Notes: $k$ represents the total sequence length, and $n$ represents the subsequence length; For all subsequences of length $k$, the goal is to change them into the minimum number of distinct sequences.
\end{itemize}

The detailed configuration of the training dataset and the function candidates in heuristics discovery are presented in Supplementary Table~\ref{tab:training-params}. 

\subsection{Evaluation Metric}
We consider two specific metrics for evaluating the performance of a SAT solver: (1) the number of SAT instances solved within the given timeout bound, and (2) the Penalized Average Runtime with a factor of 2 score (PAR-2).  Both metrics are commonly used in the SAT Competitions.


Consider a dataset of $n$ instances. Let $t_i$ be the runtime of the SAT solver on the instance $i$. The PAR-2 score is formally defined as:
\[
\text{PAR-2} = \frac{1}{n} \sum_{i=1}^{n} \tau_i
\quad
\text{where} 
\quad
\tau_i = 
\begin{cases} 
t_i, & \text{if } t_i \leq \mathcal{T}~~ \mbox{($\mathcal{T}$ is the predefined timeout bound)} \\
2\mathcal{T}, & \text{if } t_i > \mathcal{T} \text{ or the solver fails to return a result}.
\end{cases}
\]
For example, consider a benchmark dataset with three instances and a timeout bound $\mathcal{T} = 100$ seconds. The runtimes (in seconds) for the three instances are: $t_1 = 80$ for instance 1, $t_2 = 120$ for instance 2, the solver fails to return a result
for instance 3. Then
\begin{itemize}
\item since $t_1 \leq \mathcal{B}$, we have $\tau_1 = 80$;
\item
since $t_2 > \mathcal{B}$, we have $\tau_2 = 200$ (penalized);
\item
since the solver fails for instance 3, we have $\tau_3 = 200$ (penalized).
\end{itemize}
Therefore,
the PAR-2 score is given by
\[
\text{PAR-2} = \frac{1}{3} (80 + 200 + 200) = \frac{480}{3} = 160.
\]

\begin{table}[ht!]
\centering
\caption{Configuration of training set, where the indices 1 to 7 represent the following function candidates in order: rephase\_condition, rephase\_function, reduce\_condition, restart\_condition, restart\_function, varBumpActivity, claBumpActivity.}
\label{tab:training-params}
\begin{tabular}{lcc}
\toprule
Dataset & Training timeout (seconds)  & Function candidate\\
\midrule
cryptography-ascon  & 800 & $1,2,3,6$ \\
register-allocation  & 5000 & $2,3,5,6$ \\
social-golfer    & 2000 & $1,4,5,6$ \\
hashtable-safety   & 500 & $2,4,5,7$ \\
argumentation 2023      & 2000 & $1,2,3,6$ \\
argumentation 2024    & 2000 &  $1,2,3,5$ \\
hamiltonian     & 800 & $3,4,5,6$ \\
MineSweeper    & 500 & $2,4,3,7$ \\
KnightTour & 2000 & $1,3,4,7$ \\
Zamkeller   & 2000 & $1,3,4,6$ \\
EDA & 800 & $2,5,6,7$ \\

\bottomrule
\end{tabular}
\end{table}

\subsection{Search Space for Parameter Tuning}
In this paper, we adopt SMAC3 to optimize the parameters of the SAT solvers across different datasets. The search space for each solver, including the parameter name, type, description and range, are presented in  Supplementary Tables~\ref{tab:modsat-params},~\ref{tab:kissat-params}, and~\ref{tab:cadical-params}.

\begin{table}[ht!]
\centering
\caption{ModSAT configuration parameters.}
\label{tab:modsat-params}
\begin{tabular}{llp{7cm}l}
\toprule
Parameter & Type & Description & Search Space \\
\midrule
\texttt{var-decay}    & double & Variable activity decay factor & $(0, 1)$ \\
\texttt{cla-decay}    & double & Clause activity decay factor & $(0, 1)$ \\
\texttt{rnd-freq}     & double & Frequency for random variable selection & $[0, 1]$ \\
\texttt{rnd-init}     & bool   & Randomize initial activities & $\{\text{true}, \text{false}\}$ \\
\texttt{rfirst}       & int    & Base restart interval & $[1, 1e4]$ \\
\texttt{rinc}         & double & Restart interval increase factor & $(1.5, 4)$ \\
\texttt{gc-frac}      & double & Wasted memory fraction triggering garbage collection & $(0, 1)$ \\
\texttt{min-learnts} & int    & Minimum learnt clause limit & $[0, 1e6]$ \\
\bottomrule
\end{tabular}
\end{table}

\begin{table}[ht!]
\centering
\caption{Kissat configuration parameters.}
\label{tab:kissat-params}
\begin{tabular}{llp{7.5cm}l}
\toprule
Parameter & Type & Description & Search Space \\
\midrule
\texttt{chrono}        & bool   & Enable chronological backtracking & $\{0, 1\}$ \\
\texttt{eliminate}     & bool   & Enable variable elimination & $\{0, 1\}$ \\
\texttt{forcephase}    & bool   & Force initial phase assignment & $\{0, 1\}$ \\
\texttt{minimize}      & bool   & Enable clause minimization & $\{0, 1\}$ \\
\texttt{phase}         & bool   & Set initial decision phase & $\{0, 1\}$ \\
\texttt{phasesaving}   & bool   & Enable phase saving during restarts & $\{0, 1\}$ \\
\texttt{probe}         & bool   & Enable failed literal probing & $\{0, 1\}$ \\
\texttt{reduceint}     & int    & Conflict interval for clause DB reduction & $\{10^1, 10^2, 10^3, 10^4, 10^5\}$ \\
\texttt{rephaseint}    & int    & Conflict interval for phase resetting & $\{10^1, 10^2, 10^3, 10^4, 10^5\}$ \\
\texttt{restartint}    & int    & Base restart interval (conflicts) & $\{1, 10^2, 10^3, 10^4\}$ \\
\texttt{restartmargin} & int    & Rapid restart margin threshold & $\{0, 5, 10, 15, 20, 25\}$ \\
\texttt{simplify}      & bool   & Enable periodic simplification & $\{0, 1\}$ \\
\texttt{stable}        & int    & Search stability mode (0: focused, 1: stable, 2: switching) & $\{0, 1, 2\}$ \\
\texttt{target}        & int    & Target phase selection strategy (0: negative, 1: positive, 2: best) & $\{0, 1, 2\}$ \\
\texttt{tier1}         & int    & Tier 1 glue limit for learned clauses & $\{2, 3, 4, 5\}$ \\
\texttt{tier2}         & int    & Tier 2 glue limit for learned clauses & $\{6, 7, 8, 9, 10, 20, 50\}$ \\
\bottomrule
\end{tabular}
\end{table}

\begin{table}[ht!]
\centering
\caption{CaDiCaL configuration parameters.}
\label{tab:cadical-params}
\begin{tabular}{llp{7cm}l}
\toprule
Parameter & Type & Description & Search Space \\
\midrule
\texttt{chrono}       & int    & Chronological backtracking mode (0: none, 1: limited, 2: always) & $\{0, 1, 2\}$ \\
\texttt{elim}         & bool   & Enables variable elimination during simplification & $\{0, 1\}$ \\
\texttt{forcephase}   & bool   & Forces phase saving for decision variables & $\{0, 1\}$ \\
\texttt{minimize}     & bool   & Enables clause minimization during conflict analysis & $\{0, 1\}$ \\
\texttt{phase}        & bool   & Initial decision phase assignment (0: negative, 1: positive) & $\{0, 1\}$ \\
\texttt{probe}        & bool   & Enables probing (failed literal detection) & $\{0, 1\}$ \\
\texttt{reduceint}    & int    & Conflict interval for clause database reduction & $\{10^2, 10^3, 10^4, 10^5\}$ \\
\texttt{rephaseint}   & int    & Conflict interval for resetting variable phases & $\{10^1, 10^2, 10^3, 10^4, 10^5\}$ \\
\texttt{restartint}   & int    & Base restart interval (conflicts between restarts) & $\{2, 10^2, 10^3, 10^4\}$ \\
\texttt{restartmargin}& int    & Restart margin percentage (Luby sequence scaling) & $\{0, 5, 10, 15, 20, 25\}$ \\
\texttt{stabilize}    & bool   & Stabilizes search by limiting activity updates & $\{0, 1\}$ \\
\texttt{target}       & int    & Search target (0: SAT, 1: UNSAT, 2: balanced) & $\{0, 1, 2\}$ \\
\bottomrule
\end{tabular}
\end{table}

\subsection{{Comparison of Different LLMs}}

\begin{table}[ht]
\centering
\small
\caption{Context length, parameter size, and  pricing (per million tokens) for selected models.}
\label{tab:llmscomp}
\begin{tabular}{lccc}
\toprule
Model & Context length & Parameter size & Official cost ($/\!$M tokens) \\
\hline
DeepSeek-V3 & 128k & 671B & Input: \$0.27; Output: \$1.10 \\
GPT-4o &  128k & (not disclosed) & Input: \$2.50; Output: \$10.00 \\
Qwen2.5-72B & 128k & 72.7B & Input: \$1.40; Output: \$5.60 \\
Llama-3.3 & 128k & 70B & Input: \$0.49; Output: \$0.80\\
\bottomrule
\end{tabular}

\end{table}

This subsection provides a detailed comparison of different LLMs. Four representative LLMs are tested, including: DeepSeek-V3, GPT-4o, Qwen2.5-Instruct, and Llama~3.3~70B-Instruct. 

Firstly, in Supplementary  Table~\ref{tab:llmscomp} we compare their economic cost, parameters and context length.  DeepSeek-V3 and GPT-4o are accessed through their official APIs, while Qwen 2.5-72B Instruct and Llama-3.3 are invoked via third-party APIs\footnote{we access these models from https://siliconflow.cn/}. From the pricing comparison, it is evident that open-source models offer a substantially lower cost per million tokens than proprietary, closed-source models.

Specifically, GPT-4o exhibits the highest price, with input and output costs of 2.50 and 10.00 per million tokens, respectively. DeepSeek-V3 is relatively more affordable, at 0.27 for input and 1.10 for output. Other open-source models such as Qwen 2.5-72B and Llama-3.3 are  much cheaper than GPT-4o. Llama-3.3, in particular, achieves one of the lowest token costs (0.49 input, 0.80 output).

We compare the performance of different LLMs on the EDA dataset in Supplementary Table~\ref{tab:temps}. As discussed in the Method section, AutoModSAT uses a \emph{three-agent} loop: LLM coder modifies solver code; LLM evaluator identifies synonymous code; LLM repairer fixes bugs. Due to that the  LLMs evaluator and repairer focus on correctness (not diversity), we fix their temperature to $0.2$ in our framework. As for the LLMs coder,  multiple temperatures $t \in \{0.2, 0.5, 0.7, 1.0\}$ are tested. Although the theoretical temperature range is $[0, 2]$, in practice it is observed that when $t>1.0$, the probability of generating correct code drops significantly. Therefore, we limit the upper bound of coder temperature to $1.0$.

It can be seen that as the coder temperature increases, the fail rate of the generated code consistently grows, indicating that higher exploration makes the modifications more unstable. In contrast, PAR-2 improves when moving from 0.2 to intermediate temperatures and then degrades at 1.0. For DeepSeek-V3 and GPT-4o, the best PAR-2 is achieved at temperature 0.7, while Qwen-2.5 and Llama-3.3 obtain their minimum PAR-2 at 0.5. Overall, the coder performance of GPT-4o and DeepSeek-V3 is comparable and clearly stronger than that of Qwen-2.5 and Llama-3.3. The trends for the downstream evaluator and repairer agents are similar: GPT-4o attains the highest checking and repair correctness, followed by DeepSeek-V3, with Llama-3.3 and Qwen-2.5 forming a second tier. Considering both the performance and economic cost, we therefore adopt DeepSeek-V3 with coder temperature 0.7 as the default configuration in our main experiments. Note that we do not customize system prompts for all of these models. After fixing the model and temperature parameters, an  experiment with 50 iterations consumes approximately 10 million tokens.


\begin{table*}[t]
\centering
\small
\caption{\textbf{Performance vs.\ coder temperature with correctness and time breakdown on EDA.}
AutoModSAT uses a three-agent loop (coder/evaluator/repairer). We fix evaluator/repairer temperature to $0.2$ and sweep coder temperature $t\in\{0.2,0.5,0.7,1.0\}$. We report PAR-2, evaluator checkness, failure rate, repair correctness, and single call time of  the agent.}
\label{tab:temps}

\begin{tabular}{l c c c c c c c c}
\toprule
Model & \makecell{Coder \\ Temperature} &
\makecell{Coder \\ time (s)} &
\makecell{PAR-2} &
\makecell{Evaluator\\checkness (\%)} &
\makecell{Evaluator \\ time (s)} &
\makecell{Fail\\rate (\%)} &
\makecell{Repair\\correctness (\%)} &
\makecell{Repair \\ time (s)} \\
\midrule

DeepSeek-V3 & 0.2 & \multirow{4}{*}{21.05 $\pm$ 0.55} & 508.73 & 36.10 & \multirow{4}{*}{20.00 $\pm$ 0.91} & 7.85  & 68.42 & \multirow{4}{*}{20.37 $\pm$ 0.51} \\
DeepSeek-V3 & 0.5 &                                   & 389.56 & 12.34 &                                   & 10.92 & 77.03 &                                   \\
DeepSeek-V3 & 0.7 &                                   & 379.01 & 19.12 &                                   & 22.84 & 76.41 &                                   \\
DeepSeek-V3 & 1.0 &                                   & 646.18 & 6.27  &                                   & 41.63 & 82.55 &                                   \\
\midrule

GPT-4o      & 0.2 & \multirow{4}{*}{25.08 $\pm$ 1.62} & 411.22 & 55.44 & \multirow{4}{*}{22.07 $\pm$ 2.90} & 2.31  & 87.26 & \multirow{4}{*}{21.41 $\pm$ 1.54} \\
GPT-4o      & 0.5 &                                   & 394.08 & 31.67 &                                   & 11.28 & 85.91 &                                   \\
GPT-4o      & 0.7 &                                   & 378.67 & 28.54 &                                   & 13.27 & 85.12 &                                   \\
GPT-4o      & 1.0 &                                   & 449.79 & 9.83  &                                   & 32.44 & 81.73 &                                   \\
\midrule

Qwen-2.5    & 0.2 & \multirow{4}{*}{20.61 $\pm$ 0.58} & 533.95 & 44.73 & \multirow{4}{*}{18.67 $\pm$ 0.89} & 2.12  & 64.88 & \multirow{4}{*}{20.12 $\pm$ 0.74} \\
Qwen-2.5    & 0.5 &                                   & 514.37 & 30.12 &                                   & 12.83 & 73.76 &                                   \\
Qwen-2.5    & 0.7 &                                   & 521.74 & 16.41 &                                   & 25.09 & 82.34 &                                   \\
Qwen-2.5    & 1.0 &                                   & 672.66 & 7.55  &                                   & 44.38 & 79.68 &                                   \\
\midrule

Llama-3.3   & 0.2 & \multirow{4}{*}{20.55 $\pm$ 0.58} & 642.18 & 53.88 & \multirow{4}{*}{19.64 $\pm$ 0.88} & 5.67  & 53.92 & \multirow{4}{*}{19.09 $\pm$ 0.55} \\
Llama-3.3   & 0.5 &                                   & 421.03 & 29.45 &                                   & 13.77 & 81.15 &                                   \\
Llama-3.3   & 0.7 &                                   & 436.01 & 15.72 &                                   & 16.21 & 70.73 &                                   \\
Llama-3.3   & 1.0 &                                   & 685.54 & 2.83  &                                   & 35.91 & 68.54 &                                   \\
\bottomrule
\end{tabular}
\end{table*}

\subsection{Ablation Study}
Here we perform an ablation study to quantify the contribution of each component in AutoModSAT. We sequentially remove the three components: modularized SAT solver, presearch strategy, and automatic prompt optimization, to investigate their respective contributions to the performance. The results are summarized in Supplementary Table~\ref{tab:ablation}.

\begin{itemize}
    \item \textbf{AutoModSAT without Modularization.}
    Here we apply the AutoModSAT pipeline to the original, non-modularized SAT solver (i.e., MiniSat with rephasing heuristics). The most significant degradation in PAR-2 is observed among all ablations. Without modularization, LLMs struggle to generate correct and diverse heuristics, which limits the solver improvement. These results show that modularization is not merely an auxiliary enhancement but a foundational requirement for enabling effective LLM-driven solver optimization.

    \item \textbf{AutoModSAT without presearch strategy.} 
    In this variant, the LLM-based agents operate directly on the full set of function candidates without the presearch strategy. It can be seen that searching over the full space indeed leads to a slight increase in PAR-2. 

    \item \textbf{AutoModSAT without prompt optimization.}
    Here we replace the optimized coder prompts with manually written baseline prompts. Removing automatic prompt optimization consistently reduces the solver performance. This supports the common sense that high-quality proper prompts are very important for LLMs to work desirably.

\end{itemize}

Overall, the modularized SAT solver offers the largest performance gains by enabling reliable LLM-driven code generation, while presearch strategy and prompt optimization further enhance the stability and efficiency.


In Supplementary Figure~\ref{fig:search_compare}, we also report the convergence behavior of AutoModSAT with and without presearch. It clearly shows that after 50 iteration, AutoModSAT with presearch can achieve a substantially lower PAR-2 score than that without presearch. This also means that presearch can improve the computational efficiency of AutoModSAT.

\begin{figure*}[t]
    \centering
        \includegraphics[width=\textwidth]{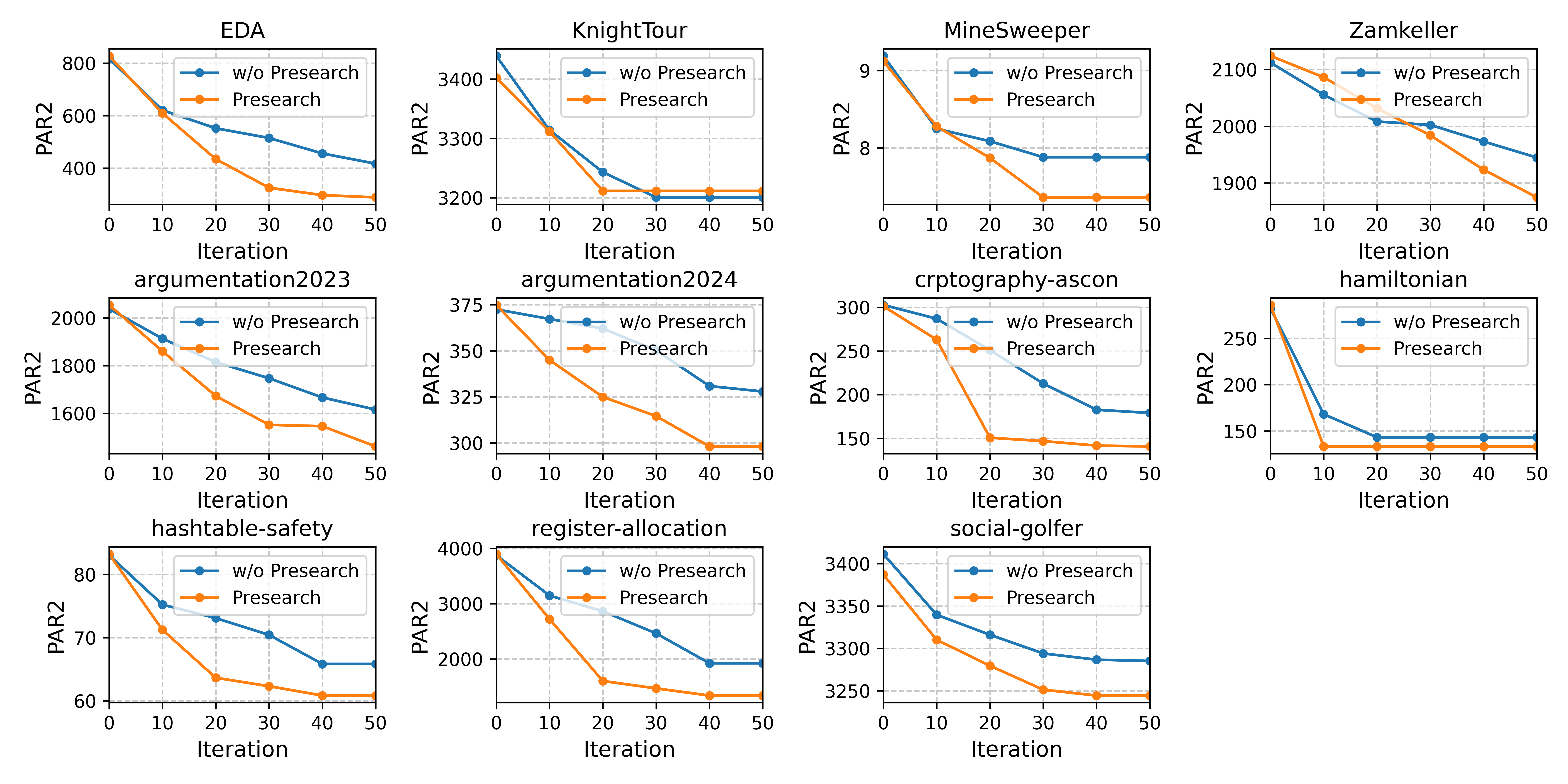}
\caption{\textbf{Convergence behavior of AutoModSAT with and without presearch}.}

\label{fig:search_compare}
\end{figure*}

\begin{table*}[t]
\centering
\caption{\textbf{Ablation Study}. Newly generated PAR-2 scores (lower is better) and solved counts 
in parentheses (higher is better) across 11 datasets for an ablation study. Bold values indicate the best PAR-2 score.}
\vskip 0.05in
\resizebox{0.9\textwidth}{!}{
\begin{tabular}{lccccc}
\toprule
Dataset & ModSAT & 
\makecell{AutoModSAT \\ w/o modularization}  & \makecell{AutoModSAT \\ w/o presearch} & \makecell{AutoModSAT \\ w/o prompt optimization} & AutoModSAT \\
\midrule
cryptography-ascon     
& 208.60 (20) & 200.45 (20) & 178.30 (20) & 180.12 (20) & \textbf{140.72 (20)} \\

register-allocation 
& 7213.23 (6) & 5120.10 (10) & 1920.55 (17) 
& 3055.32 (14) & \textbf{1177.80 (18)} \\

social-golfer 
& 7836.30 (5) & 7602.11 (5) & 7420.24 (6) & 7588.92 (5) & \textbf{7265.45 (6)} \\

hashtable-safety 
& 83.25 (20) & 80.14 (20) & 67.92 (20) & 78.50 (20) & \textbf{60.82 (20)} \\

argumentation 2023 
& 4695.25 (13) & 4002.33 (13) & 3310.25 (14) & 3535.10 (14) & \textbf{3229.65 (14)} \\

argumentation 2024 
& 2745.57 (17) & 780.22 (20) & 305.70 (21)
& 465.41 (21) & \textbf{232.14 (21)} \\

hamiltonian 
& 577.48 (38) & 355.20 (39) & 162.90 (40) & 310.48 (39) & \textbf{133.13 (40)} \\
\midrule

MineSweeper 
& 9.12 (88) & 9.06 (88) & 7.88 (88) & 9.90 (88) & \textbf{7.36 (88)} \\

KnightTour 
& 8769.77 (7) & 8620.22 (8) & 8088.31 (11) & 8120.50 (11) & \textbf{7815.68 (13)} \\

Zamkeller 
& 3485.35 (54) & 2622.10 (63) & 2490.44 (62) & 2455.80 (64) & \textbf{2053.21 (66)} \\
\midrule

EDA 
& 825.36 (48) & 930.55 (48) & 510.12 (49) & 722.44 (48) & \textbf{376.01 (49)} \\
\bottomrule
\end{tabular}}
\label{tab:ablation}
\end{table*}

\subsection{Generalization Tests}

\paragraph{In-domain generalization tests.} To further distinguish domain specialization from structural overfitting, we conduct the generalization tests where the training and test data share the same domain. The new test data are generated as follows. First, there are three families that have appeared repeatedly in the SAT Competitions and thus contain additional instances\footnote{Note that the other four families do not have additional instances.} beyond the ones used (SAT Competitions 2023 and 2024) in the solver optimization: argumentation (156 instances), hamiltonian (1060 instances), and social-golfer (19 instances). 
For argumentation and hamiltonian, we randomly sample 100 previously unseen instances as the test data, while for social-golfer we use all 19 instances. 
Second, for the two families generated by Picat (MineSweeper, KnightTour), we sample extra 88 and 56 instances per family as the test data.

The generalization test results are shown in Supplementary Table~\ref{tab:gene_inside}. 
It can be observed that the heuristics discovered on argumentation, hamiltonian and social-golfer remain consistently competitive on their in-domain test data, and the same trend can be observed for the Picat-generated families. This indicates that as an automation framework of domain-specific solvers, the optimization by AutoModSAT does not merely memorize the training data, but can generalize well on the test data in the same domain.

\begin{table*}[t]
\centering
\belowrulesep=0pt
\aboverulesep=0pt
\caption{\textbf{In-domain generalization tests.}
This table shows  PAR-2 (solved instances in parentheses) of the produced solvers tested on the new in-domain data.  Note that the argumentation dataset uses solvers optimized on argumentation 2023. Bold values indicate the best PAR-2 score.}
\label{tab:gene_inside}
\resizebox{\textwidth}{!}{
\begin{tabular}{lcccccccc}
\toprule
Dataset & MiniSat & ModSAT & ModSAT para & Kissat & Kissat para & CaDiCaL & CaDiCaL para & AutoModSAT \\
\midrule
social-golfer   & 5100.23 (10) & 4805.77 (11) & 4552.61 (12) & 4321.08 (13) & 4288.90 (13) & 4269.09 (13) & 4154.32 (13) & \textbf{4067.68 (14)} \\
argumentation   & 8750.66 (30) & 6420.14 (41) & 6012.45 (44) & 5755.32 (45) & 5488.76 (48) & 5597.02 (47) & 5324.11 (50) & \textbf{5083.01 (54)} \\
hamiltonian     &  \,980.44 (90) &  \,820.15 (92) &  \,690.73 (94) &  \,640.20 (95) &  \,410.55 (97) &  \,560.34 (96) &  \,245.67 (98) & \textbf{118.39 (100)} \\
MineSweeper     & \,10.42 (88) & \,9.31 (88)  & \,8.07 (88)  & 148.76 (88)  & \,49.28 (88)  & \,56.14 (88)  & \,44.02 (88)  & \textbf{7.51 (88)} \\
KnightTour      & 9198.35 (5)  & 8794.12 (7)  & 8127.46 (11) & 8771.08 (7)  & 8599.41 (8)  & 8386.27 (9)  & 8310.58 (10) & \textbf{7864.93 (13)} \\
\bottomrule
\end{tabular}
}
\end{table*}

\paragraph{Cross-domain generalization tests.} Further generalization tests on the cross-domain data have been conducted and results are presented in Supplementary Table~\ref{tab:par2_by_dataset_part1} and  Supplementary Table~\ref{tab:par2_by_dataset_part2}.
Overall, there are very few {heuristics} which can generalize cross the domain, which is natural given that the datasets differ from each other substantially and the optimization is conducted on particular dataset. Nonetheless, certain heuristics exhibit broader applicability: the heuristic discovered in hamiltonian achieves consistently competitive performance across multiple other datasets, indicating that some discovered strategies encode reusable search behaviors beyond their source domain.

\begin{table*}[t]
\centering
\scriptsize
\belowrulesep=0pt
\aboverulesep=0pt
\caption{\textbf{Cross-domain generalization tests (part 1).} This table shows PAR-2 of the solver discovered from one dataset but evaluated on the other datasets (row: source dataset, column: evaluation dataset).
}

\label{tab:par2_by_dataset_part1}
\resizebox{\textwidth}{!}{
\begin{tabular}{lrrrrr}
\toprule
Dataset  & cryptography-ascon & register-allocation & social-golfer & hashtable-safety & argumentation 2023 \\
\midrule
cryptography-ascon   & 140.72 (20) & 10000.00 (0) & 9078.14 (2) & 566.76 (20) & 7618.30 (6)\\
register-allocation  & 591.36 (20)  & 1177.80 (18) & 10000.00 (0) & 175.23 (20) & 3734.85 (14)\\
social-golfer        & 2770.50 (15) & 9646.80 (1) & 7265.45 (6) & 100.85 (20)  & 4199.48 (13)\\
hashtable-safety     & 442.36 (20) & 9621.51 (1) & 10000.00 (0) & 60.82 (20)  & 3472.21 (14)\\
argumentation 2023   & 238.68 (20) & 8655.26 (3) & 10000.00 (0) & 135.31 (20) & 3229.65 (14)\\
argumentation 2024   & 2857.93 (15) & 8627.86 (3) & 10000.00 (0) & 6118.72 (10) & 3273.99 (14)\\
hamiltonian          & 693.21 (20) & 9040.47 (2) & 7160.60 (6) & 567.41 (20) & 3434.80 (14)\\
MineSweeper          & 352.43 (20) & 8674.13 (3) & 10000.00 (0) & 196.22 (20) & 4181.09 (13)\\
KnightTour           & 691.36 (20) & 8823.31 (3) & 10000.00 (0) & 215.33 (20) & 3483.80 (14)\\
Zamkeller            & 339.15 (20) & 9197.00 (2) & 10000.00 (0) & 67.09 (20)  & 8585.55 (4)\\
EDA                  & 694.26 (20) & 9602.40 (1) & 10000.00 (0) & 233.31 (20) & 4281.80 (13)\\
\bottomrule
\end{tabular}
}
\end{table*}
\begin{table*}[t]
\centering
\scriptsize
\belowrulesep=0pt
\aboverulesep=0pt
\caption{\textbf{Cross-domain generalization tests (part 2).} This table shows PAR-2 of the solver discovered from one dataset but evaluated on the other datasets (row: source dataset, column: evaluation dataset).
}
\label{tab:par2_by_dataset_part2}
\resizebox{\textwidth}{!}{
\begin{tabular}{lrrrrrr}
\toprule
Dataset  & argumentation 2024 & hamiltonian & MineSweeper & KnightTour & Zamkeller & EDA \\
\midrule
cryptography-ascon   & 5451.93 (12) & 2406.25 (35) & 7.59 (88) & 8624.27 (8) & 7597.07 (22) & 749.19 (47)\\
register-allocation  & 873.64 (21) & 657.01 (39)  & 8.33 (88) & 8290.95 (10) & 5792.07 (37) & 809.75 (47)\\
social-golfer        & 1733.74 (19) & 267.90 (40) & 7.51 (88) & 8396.78 (9) & 6567.69 (30) & 609.15 (48) \\
hashtable-safety     & 516.20 (21) & 260.45 (40) & 7.62 (88) & 8590.95 (8) & 7721.79 (23) & 557.97 (48)\\
argumentation 2023   & 379.91 (21) & 253.03 (40) & 8.44 (88) & 8588.50 (8) & 8000.07 (21) & 405.91 (48)\\
argumentation 2024   & 232.14 (21) & 768.64 (39) & 10.45 (88) & 8098.60 (11) & 7969.48 (18) & 2269.94 (41) \\
hamiltonian          & 863.64 (21) & 133.13 (40) & 8.49 (88) & 8280.72 (10) & 4711.09 (47) & 2757.74 (38) \\
MineSweeper          & 1073.23 (20) & 257.87 (40) & 7.36 (88) & 8590.95 (8) & 7594.35 (21) & 839.55 (47) \\
KnightTour           & 637.32 (21) & 227.11 (40) & 8.22 (88) & 7815.68 (13) & 7592.07 (21)& 1022.73 (47)\\
Zamkeller            & 7589.31 (6) & 4219.22 (29) & 8.14 (88) & 8153.99 (11) & 2053.21 (66) & 388.61 (49)\\
EDA                  & 988.32 (20) & 239.86 (40) & 8.30 (88) & 8590.95 (8) & 4397.11 (49) & 376.01 (49) \\
\bottomrule
\end{tabular}
}
\end{table*}

\subsection{{Correctness  Verification}}
\label{sec:correctness}

After discovering heuristics in a SAT solver, we explicitly validate the correctness of the generated solvers over  all the problem instances.

\begin{itemize}
    \item \textbf{SAT case validation.}
    For every instance reported as satisfiable, the solver outputs a full assignment to all relevant variables. 
    We then independently validate this assignment by substituting it back into the original CNF formula to see whether every clause is satisfied. 
    This checker is implemented as a separate, simple CNF evaluator, and
    any instance for which the reported assignment does not satisfy the CNF is flagged as incorrect and the corresponding solver variant is rejected by AutoModSAT.
    
    \item \textbf{UNSAT proof verification.} 
    For every instance reported as unsatisfiable, the solver is required to emit a DRAT proof. 
    We verify this proof using the external proof checker drat-trim~\cite{drat-trim}, which independently validates that the DRAT sequence constitutes a correct refutation of the original CNF. If drat-trim rejects the proof, the UNSAT result is deemed invalid and the corresponding solver mutation is rejected by AutoModSAT.
    
    \item \textbf{Cross solver consistency.} 
    Finally, for each dataset, we compare the SAT/UNSAT status reported by all tested solvers (baseline solvers and AutoModSAT-discovered solvers). In our experiments, no inconsistencies in SAT/UNSAT status were observed across the solvers, and all reported SAT assignments and UNSAT proofs successfully passed the above validation pipeline.

\end{itemize}

\section{Analysis of Discovered Heuristics}
In this section, to demonstrate LLMs' ability of generating effective heuristics, we provide more contrastive examples (one for each function candidate), along with the explanations for the changes.

\subsection{Examples of Discovered Heuristics}
Supplementary Figure~\ref{fig:claBumpActivity} provides an example for the updated claBumpActivity function, which introduces two key enhancements in contrast to the original implementation. First, during activity rescaling, it enforces a minimum activity threshold (min\_activity = 1e-20) to prevent clauses from becoming numerically insignificant after scaling. This preserves the relevance of historically important clauses and avoids premature elimination from the learning process. Second, it incorporates dynamic decay adjustment based on recent conflict rates: when conflicts exceed 1000 and the LBD queue surpasses 50 entries, cla\_inc is scaled down proportionally to the conflict density (with a floor of $0.8$). This procedure adaptively moderates the activity growth during high-conflict phases, prioritizing recent high-impact clauses while maintaining stability. Together, these refinements yield a more balanced clause management strategy which utilizes valuable learned clauses while dynamically optimizes activity decay for solver efficiency.

Supplementary Figure~\ref{fig:varBumpActivity} provides an example of the updated varBumpActivity function, which introduces three key improvements over the original. First, it scales the increment by (1.0 + 0.1 * decisionLevel()), prioritizing variables involved in recent decisions to accelerate conflict-driven learning. Second, the rescaling mechanism uses a larger threshold (1e100) and finer scale factor (1e-100), while preserving variable relevance by enforcing a minimum activity floor (1e-100) to maintain relative ordering and prevent premature underflow. Third, it optimizes heap management through conditional updates: dynamically adjusting the variable's heap position only when its activity exceeds the current maximum, or inserting undefined decision variables lazily. These enhancements collectively improve search guidance, reduce floating-point stability issues, and minimize unnecessary data structure operations.

Supplementary Figure~\ref{fig:restartcondition} provides an example of the updated restart\_condition function. It improves upon the original one by replacing the static threshold approach with a dynamic, performance-driven restart strategy that adapts to real-time solver behavior. Instead of relying on fixed queue sizes and hard coded multipliers, the new version  calculates the restart thresholds using multiple runtime metrics: it combines average LBD (measuring clause quality) with conflict rates (tracking solver progress) to dynamically adjust the restart timing based on problem difficulty. Crucially, it introduces a progress-sensitive mechanism that aggressively lowers thresholds when stagnation is detected (progressEstimate changes < 0.01), enabling proactive recovery from plateaus, a capability absent in the original. This multi-factor approach yields more precise restart decisions, reduces wasteful computations, and enhances solver adaptability across diverse SAT instances while maintaining robustness during initialization through default thresholds.

Supplementary Figure~\ref{fig:restartfunction} provides an example of the updated restart\_function. Unlike the original version which always resets to decision level 0 (a full restart), the enhanced function dynamically calculates two exponential moving averages of conflict difficulty (fast\_avg and slow\_avg) using Literal Block Distance (LBD) scores. By analyzing the ratio between these averages, it intelligently selects one of three restart depths: full restart (level 0) for deteriorating conflict quality, partial restart (mid-level) for moderately harder conflicts, or minimal restart (current level -1) for stable conditions. This adaptability preserves useful learned clauses during partial/minimal restarts and reduces redundant recomputation. Additionally, periodic clause database reduction (every 16 restarts) curbs memory growth, while rebuilding the variable order heap ensures that the branching decisions can reflect the updated activity scores. 

Supplementary Figure~\ref{fig:rephasecondition} provides an example of the updated rephase\_condition function. In contrast to the prior version which solely relies on a fixed rephase limit, the new implementation dynamically adjusts rephasing intervals based on real-time search progress and conflict density. By calculating normalized progress through trail size changes and setting variable-driven thresholds (e.g., $2\%$ of total variables), it detects stagnation when progress falls below expectation and then reduces the subsequent rephase intervals exponentially. On the other hand, substantial progress triggers gradual interval expansion. This self-tuning capability optimizes computational efficiency: it minimizes unnecessary rephasing during the search loop, thereby improving solution convergence.

Supplementary Figure~\ref{fig:rephasefunction} provides an example of the updated rephase\_function. First, it implements dynamic rephase limit adjustment by scaling rephase\_limit based on progress measured through conflict resolution (conflictR). If progress occurs, the limit increases by $50\%$ to exploit productive phases more aggressively; otherwise, it decays by $10\%$ (with a lower bound of 512) to conserve resources during stagnation. This replaces the original static increment (+= 8192) and fixed decay (threshold *= 0.9), enabling context-sensitive resource allocation. Second, the refined phase selection strategy uses weighted probabilities with four distinct policies instead of the original rigid three-policy cascade: local-best phases ($40\%$), global phase inversion ($30\%$), randomized phases for low-activity variables ($20\%$), and user-specified phases ($10\%$). In addition, the introduction of randomization for less-active variables helps escape local optima. Finally, adaptive threshold dynamically scales with the solver’s state (threshold = trail.size() * 0.8).

Supplementary Figure~\ref{fig:reducecondition} provides an example of the updated reduce\_condition function, which enhances the original version through four key improvements. First, it retains the core check for absolute learnt clause limits (learnts.size() >= max\_learnts), ensuring baseline constraint adherence. Second, it introduces memory pressure awareness by triggering reduction when wasted clause memory exceeds $80\%$ of the garbage collection threshold (ca.wasted() > ca.size() * garbage\_frac * 0.8). This proactively mitigates memory bloat and improves cache efficiency. Third, a learnt-to-original clause ratio check (learnts.size() > 2 * nClauses()) prevents learnt clauses from disproportionately dominating the formula, maintaining balanced reasoning. Finally, a conflict-driven heuristic (conflictR > 1000 \&\& learnts.size() > max\_learnts * 0.8) adapts to high-conflict phases by initiating earlier reduction, thus accelerating recovery from solver stagnation. These  criteria boost robustness: they minimize redundant computation through memory-sensitive garbage collection, preserve clause quality via ratio controls, and dynamically respond to runtime behavior.

\subsection{Novelty Assessment of Discovered Heuristics}

To address questions regarding the fundamental novelty of the heuristics discovered by AutoModSAT, we conduct a systematic investigation. It should be noted  that an evaluation of the true novelty is challenging due to the infeasibility of exhaustively enumerating all open-source codes. Our goal is therefore to perform a rigorous comparison against a well-defined corpus of modern SAT solvers. The assessment process is comprised of three key steps:

\begin{itemize}
    \item \textbf{Solver collection.} We collect the top-10 solvers from major track in SAT competitions between 2020 and 2025 to establish a representative benchmark of state-of-the-art techniques.
    
    \item \textbf{Heuristic extraction.} For each solver, we identify its source files with an LLM to parse the code, identify heuristic rules, and normalize their descriptions into a consistent format.
    
    \item \textbf{Automated comparison.} We adopt an LLM-as-a-judge strategy to automatically compare the heuristics discovered by AutoModSAT with those extracted from the reference corpus. The comparison follows a binary (True/False) format, evaluating whether a given heuristic shares similar ideas or mechanisms with any in the corpus. Heuristics marked as ``similar ideas'' are then subject to manual verification by human experts to confirm or refute conceptual overlap.

\end{itemize}
\begin{table}[htbp]
\centering
\caption{Solvers which share some similar ideas with the discovered heuristics.}
\label{tab:share}
\begin{tabular}{cc}
\toprule
\multicolumn{1}{c}{Heuristics} & \multicolumn{1}{c}{Solvers with similar idea}  \\
\midrule
\multirow{1}{*}{claBumpActivity} & Maple\_MBDR\_Cent\_PERM  \\
\cmidrule(lr){1-2}
\multirow{3}{*}{reduce\_condition} & MapleLCMDistChrBt-DL \\
 & kissat-sc2022-bulk  \\
 & kissat\_inc  \\
\cmidrule(lr){1-2}
\multirow{6}{*}{rephase\_condition} & kissat-sc2024  \\
 & BreakID-kissat  \\
 & Kissat-CURE  \\
 & kissat-sat\_crvr\_gt  \\
 & Kissat\_MAB\_prop  \\
 & Kissat\_MAB-DC  \\
\cmidrule(lr){1-2}
\multirow{3}{*}{rephase\_function} & kissat-sc2020-default  \\
 & CaDiCal-PriPro  \\
 & kissat-pred  \\
\cmidrule(lr){1-2}
\multirow{4}{*}{restart\_condition} & Maple\_simp21  \\
 & Maple\_MBDR\_Cent\_PERM  \\
 & Maple\_MBDR\_BJL7\_Loc  \\
 & Maple\_MBDR\_BJL6\_Tie  \\
\cmidrule(lr){1-2}
\multirow{5}{*}{restart\_function} & kissat-sc2025  \\
 & Cadical-PriPro  \\
 & kissat-1.0.3  \\
 & Maple\_MBDR\_BJL7\_Loc  \\
 & kissat-pred  \\
\cmidrule(lr){1-2}
\multirow{4}{*}{varBumpActivity} & Maple\_MBDR\_Cent\_PERM  \\
 & kissat-1.0.3  \\
 & kissat-pred  \\
 & CaDiCal  \\
\bottomrule
\end{tabular}
\end{table}
Finally, we have obtained 60 solver repositories and sorted out the seven heuristic categories  from each solver into structured files. After running the automated comparison, we find out that some solvers that potentially share similar ideas with the discovered heuristics by AutoModSAT, which is shown in Supplementary Table~\ref{tab:share}. However, further manual verification confirms that none of these heuristics exhibit completely the same implementations.

Notably, compared with the original heuristics, the heuristics discovered by AutoModSAT tend to integrate multiple rules, resulting in more complex and hybrid heuristics. While in a few instances, it is likely  that some rules draw inspiration from prior solvers. For example, the average LBD computation used in restart conditions appears in the Maple family of solvers. However, other key elements such as the incorporation of a \texttt{progressEstimate} factor do not appear in any existing implementations. Overall, these findings suggest that there are new heuristics discovered by AutoModSAT.


\begin{figure}
\begin{cppcode}{original claBumpActivity}
inline void Solver::claBumpActivity (Clause& c) {
        if ( (c.activity() += cla_inc) > 1e20 ) {
            // Rescale:
            for (int i = 0; i < learnts.size(); i++)
                ca[learnts[i]].activity() *= 1e-20;
            cla_inc *= 1e-20; } 
        }
\end{cppcode}

\begin{cppcode}{discovered claBumpActivity}
inline void Solver::claBumpActivity (Clause& c) {
    // Adaptive scaling based on clause length and conflict participation
    if ((c.activity() += cla_inc) > 1e20) {
        // Maintain minimum activity while scaling to preserve relevance
        double scale_factor = 1e-20;
        double min_activity = 1e-20;
        for (int i = 0; i < learnts.size(); i++) {
            ca[learnts[i]].activity() *= scale_factor;
            if (ca[learnts[i]].activity() < min_activity)
                ca[learnts[i]].activity() = min_activity;
        }
        cla_inc *= scale_factor;
        
        // Stabilize scaling by maintaining lower bound on clause activity
        if (cla_inc < min_activity)
            cla_inc = min_activity;
    }
    // Dynamic decay adjustment based on recent conflict rate
    if (conflicts > 1000 && lbd_queue_size > 50) {
        double conflict_scale = 1.0 - 0.01 * (lbd_queue_size / 50.0);
        cla_inc *= conflict_scale > 0.8 ? conflict_scale : 0.8;
    }
}
\end{cppcode}
    \caption{Example of LLM discovered heuristics of claBumpActivity}
    \label{fig:claBumpActivity}
\end{figure}

\begin{figure}
\begin{cppcode}{original varBumpActivity}
inline void Solver::varBumpActivity(Var v, double inc) {
    activity[v] += inc;
    // Rescale condition to prevent floating-point overflow earlier
    if (activity[v] > 1e50) {
        for (int i = 0; i < nVars(); i++)
            activity[i] *= 1e-50;
        var_inc *= 1e-50;
        }
    }
\end{cppcode}

\begin{cppcode}{discoverd varBumpActivity}
inline void Solver::varBumpActivity(Var v, double inc) {
    // Scale the increment based on decision level to prioritize recent variables
    double scaled_inc = inc * (1.0 + 0.1 * decisionLevel());
    activity[v] += scaled_inc;
    // More gradual rescaling that preserves relative ordering better
    if (activity[v] > 1e100) {
        double scale_factor = 1e-100;
        for (int i = 0; i < nVars(); i++) {
            activity[i] *= scale_factor;
            // Prevent variables from becoming completely inactive
            if (activity[i] < 1e-100) activity[i] = 1e-100;
        }
        var_inc *= scale_factor;
    }
    // Optimized heap update with early exit for common case
    if (order_heap.inHeap(v)) {
        if (activity[v] > activity[order_heap[0]]) {
            order_heap.decrease(v);
        }
    } else if (decision[v] && value(v) == l_Undef) {
        order_heap.insert(v);
    }
}
\end{cppcode}
    \caption{Example of LLM discovered heuristics of varBumpActivity}
    \label{fig:varBumpActivity}
\end{figure}

\begin{figure}
\begin{cppcode}{original restart\_condition}
bool Solver::restart_condition(){
    if ((lbd_queue_size == 50 && 0.8 * fast_lbd_sum / lbd_queue_size > slow_lbd_sum / conflicts) || !withinBudget())
        return true;
    else
        return false;

}
\end{cppcode}

\begin{cppcode}{discoverd restart\_condition}
bool Solver::restart_condition() {
    // Base restart condition on conflict count
    if (conflicts <= 0) return false;
    
    // Dynamic restart threshold based on recent solver performance
    double restart_threshold;
    
    if (lbd_queue_size > 0) {
        // Use both average LBD and conflict rate for adaptive restart
        double avg_lbd = fast_lbd_sum / lbd_queue_size;
        double conflict_rate = (double)conflictR / (double)conflicts;
        
        // Adjust threshold based on problem difficulty (higher LBD = harder problem)
        restart_threshold = restart_first * (0.8 + 0.4 * avg_lbd) * (1.0 + 0.5 * conflict_rate);
        
        // More aggressive restart if we're making little progress
        if (progressEstimate() - last_rephase_progress < 0.01) {
            restart_threshold *= 0.7;
        }
    } else {
        // Default threshold for initial phase
        restart_threshold = restart_first;
    }
    
    // Check if we've reached the threshold
    if (conflictR >= restart_threshold) {
        conflictR = 0;
        return true;
    }   
    return false;
}
\end{cppcode}
    \caption{Example of LLM discovered heuristics of restart condition}
    \label{fig:restartcondition}
\end{figure}

\begin{figure}
\begin{cppcode}{original restart\_function}
void Solver::restart_function(){
    fast_lbd_sum = lbd_queue_size = lbd_queue_pos = 0;
    int level = 0;
    cancelUntil(level);
}
\end{cppcode}

\begin{cppcode}{discovered restart\_function}
void Solver::restart_function() {

    // Update restart statistics
    static double fast_avg = 0, slow_avg = 0;
    static int restart_count = 0;
    
    // Calculate moving averages of LBD scores
    if (lbd_queue_size > 0) {
        double curr_fast = fast_lbd_sum / lbd_queue_size;
        fast_avg = 0.9 * fast_avg + 0.1 * curr_fast;
        slow_avg = 0.99 * slow_avg + 0.01 * curr_fast;
    }

    // Determine restart level based on recent performance
    int restart_level = 0;
    if (fast_avg > 0 && slow_avg > 0) {
        double ratio = fast_avg / slow_avg;
        if (ratio > 1.2) {
            // Recent conflicts are harder - do a full restart
            restart_level = 0;
        } else if (ratio > 1.0) {
            // Partial restart - keep some learned clauses
            restart_level = std::max(0, decisionLevel() / 2);
        } else {
            // Recent conflicts are easier - minimal restart
            restart_level = std::max(0, decisionLevel() - 1);
        }
    }

    // Reset LBD statistics
    fast_lbd_sum = lbd_queue_size = lbd_queue_pos = 0;
    
    // Perform the restart
    cancelUntil(restart_level);
    
    // Occasionally reduce the clause database after restarts
    if (restart_count++ \% 16 == 15) {
        reduceDB();
    }

    // Rebuild the order heap to reflect any activity changes
    rebuildOrderHeap();
}
\end{cppcode}
    \caption{Example of LLM discovered heuristics of restart function}
    \label{fig:restartfunction}
\end{figure}

\begin{figure}[ht!]
\begin{cppcode}{original rephase\_condition}
bool Solver::rephase_condition() {
    if (rephases >= rephase_limit) return true;
    else return false;
}
\end{cppcode}

\begin{cppcode}{discovered rephase\_condition}
bool Solver::rephase_condition() {
    // Adaptive rephasing based on progress and conflict density
    const int base_rephase_limit = 1024;
    const double progress_factor = 0.02;  // 2% of total variables
    
    if (conflictR < rephase_limit) 
        return false;

    // Calculate normalized progress since last rephase
    int progress = trail.size() - last_rephase_progress;
    int progress_threshold = std::max(50, (int)(nVars() * progress_factor));
    
    // Dynamic limit adjustment based on search progress
    if (progress < progress_threshold) {
        // Stagnation detected - reduce next rephase interval
        rephase_limit = std::max(base_rephase_limit, rephase_limit * 2/3);
    } else {
        // Good progress - expand next rephase interval
        rephase_limit = std::min(base_rephase_limit * 16, rephase_limit * 3/2);
    }

    // Reset conflict counter and update progress tracker in rephase_function
    return true;
}
\end{cppcode}
    \caption{Example of LLM discovered heuristics of rephase condition}
    \label{fig:rephasecondition}
\end{figure}

\begin{figure}[ht!]
\begin{cppcode}{original rephase\_function}
void Solver::rephase_function() {
    int var_nums = nVars();
    conflictR = 0, rephases = 0, threshold *= 0.9, rephase_limit += 8192;Add commentMore actions
    int phase_rand = rand() \% 100;      
    if ((phase_rand -= 40) < 0){
        for (int i = 0; i < var_nums; i++){
            polarity[i] = local_best[i];
        }
    }
    else if ((phase_rand -= 25) < 0){
        for (int i = 0; i < var_nums; i++){
            polarity[i] = 1-local_best[i];
        }
    }
    else if ((phase_rand -= 15) < 0){
        for (int i = 0; i < var_nums; i++){
            polarity[i] = !polarity[i];
        }
    }
    else {
        for (int i = 0; i < var_nums; i++){
            polarity[i] = saved[i];
        }
    }
}
\end{cppcode}
    \label{fig:rephasefunction_original}
\end{figure}

\begin{figure}[ht!]
\begin{cppcode}{discovered rephase\_function}
void Solver::rephase_function() {
    // Adaptive rephase limit adjustment
    if (rephases > 0 && conflictR > last_rephase_progress) {
        // Progress was made - be more aggressive with rephasing
        rephase_limit = rephase_limit * 1.5;
    } else {
        // No progress - be more conservative
        rephase_limit = rephase_limit * 0.9;
        if (rephase_limit < 512) rephase_limit = 512;
    }
    last_rephase_progress = conflictR;
    rephase_count++;

    // Dynamic phase selection strategy
    double rand_val = drand(random_seed);
    if (rand_val < 0.4) {
        // 40\% chance: Use local best phases
        for (int v = 0; v < nVars(); v++) {
            polarity[v] = local_best[v];
        }
        if (verbosity >= 1) printf("| Rephase: Local best phases |\n");
    } else if (rand_val < 0.7) {
        // 30\% chance: Invert all phases
        for (int v = 0; v < nVars(); v++) {
            polarity[v] = !polarity[v];
        }
        if (verbosity >= 1) printf("| Rephase: Inverted all phases |\n");
    } else if (rand_val < 0.9) {
        // 20\% chance: Random phases for variables with low activity
        double activity_threshold = 0.2 * var_inc;
        for (int v = 0; v < nVars(); v++) {
            if (activity[v] < activity_threshold) {
                polarity[v] = drand(random_seed) < 0.5;
            }
        }
        if (verbosity >= 1) printf("| Rephase: Randomized low-activity phases |\n");
    } else {
        // 10\% chance: Original user phases
        for (int v = 0; v < nVars(); v++) {
            if (user_pol[v] != l_Undef) {
                polarity[v] = (user_pol[v] == l_True);
            }
        }Add commentMore actions
        if (verbosity >= 1) printf("| Rephase: Reset to user phases |\n");
    }

    // Reset the threshold for next local best phase tracking
    threshold = trail.size() * 0.8;
    cancelUntil(0);
}
\end{cppcode}
    \caption{Example of LLM discovered heuristics of rephase function}
    \label{fig:rephasefunction}
\end{figure}

\begin{figure}[ht!]
\begin{cppcode}{original reduce\_condition}
bool Solver::reduce_condition()
{   
    if (learnts.size() >= max_learnts)  return true;
    else return false;
}
\end{cppcode}

\begin{cppcode}{discovered reduce\_condition}
bool Solver::reduce_condition()
{
    // Check if we've reached the absolute limit of learnt clausesAdd commentMore actions
    if (learnts.size() >= max_learnts) return true;
    
    // Consider memory pressure and garbage collection needs
    if (ca.wasted() > ca.size() * garbage_frac * 0.8) return true;
    
    // Consider the ratio between learnt and original clauses
    if (learnts.size() > 0 && learnts.size() > 2 * nClauses()) return true;
    
    // Consider recent solver performance (conflict rate)
    if (conflictR > 1000 && learnts.size() > max_learnts * 0.8) return true;
    
    return false;
}
\end{cppcode}
    \caption{Example of LLM discovered heuristics of reduce condition}
    \label{fig:reducecondition}
\end{figure}

\end{document}